\documentclass[preprint,5p, times, twocolumn, authoryear]{elsarticle}

\usepackage{amssymb}

\usepackage{amsmath,amsfonts}
\usepackage{array}
\usepackage{textcomp}
\usepackage{stfloats}
\usepackage{url}
\usepackage{verbatim}
\usepackage{graphicx}
\usepackage{cite}
\usepackage{breqn}

\usepackage{enumitem}
\setlist[description]{font=\normalfont\itshape\space}
\usepackage{amsmath, bm}
\usepackage{amsfonts}
\usepackage{dsfont}
\usepackage[section]{placeins}
\usepackage{subcaption,graphicx}
\usepackage{mathrsfs}
\usepackage{float}
\usepackage[ruled,vlined,linesnumbered,lined,boxed,commentsnumbered]{algorithm2e}
\usepackage[utf8]{inputenc} 
\usepackage[T1]{fontenc}    
\usepackage[colorlinks=true,linkcolor=blue, citecolor=blue]{hyperref}       
\usepackage{url}            
\usepackage{booktabs}       
\usepackage{amsfonts, bm, amsmath}       
\usepackage{thmtools, thm-restate}
\usepackage{amssymb}
\usepackage{nicefrac}       
\usepackage{xcolor}         

\usepackage{dsfont}
\usepackage{mathrsfs}
\usepackage{placeins}
\usepackage{mathtools}
\usepackage{float}
\usepackage{graphicx}
\usepackage{adjustbox}
\usepackage[capitalize,noabbrev]{cleveref}

\newcommand{\E}{\mathbb{E}} 
 
\newcommand{\N}[2]{{\mathcal{N}\left(#1, #2\right)}}

\newtheorem{definition}{Definition}

\usepackage{float}
\makeatletter
\def\ps@pprintTitle{%
 \let\@oddhead\@empty
 \let\@evenhead\@empty
 \def\@oddfoot{}%
 \let\@evenfoot\@oddfoot}
\makeatother

\usepackage[round]{natbib}
\begin{document}
\begin{frontmatter}

\title{How to Evaluate Uncertainty Estimates in Machine Learning for Regression?}

\author{Laurens Sluijterman}
\address{Department of Mathematics, Radboud University, \\P.O. Box 9010-59, 6500 GL, Nijmegen, Netherlands}
\ead{L.Sluijterman@math.ru.nl}
\author{Eric Cator}
\address{Department of Mathematics, Radboud University}
\ead{e.cator@science.ru.nl}
\author{Tom Heskes}
\address{Institute for Computing and Information Sciences, Radboud University}
\ead{Tom.Heskes@ru.nl}

\begin{abstract}
As neural networks become more popular, the need for accompanying uncertainty estimates increases. There are currently two main approaches to test the quality of these estimates. Most methods output a density. They can be compared by evaluating their loglikelihood on a test set. Other methods output a prediction interval directly. These methods are often tested by examining the fraction of test points that fall inside the corresponding prediction intervals. Intuitively both approaches seem logical. However, we demonstrate through both theoretical arguments and simulations that both ways of evaluating the quality of uncertainty estimates have serious flaws. Firstly, both approaches cannot disentangle the separate components that jointly create the predictive uncertainty, making it difficult to evaluate the quality of the estimates of these components. Secondly, a better loglikelihood does not guarantee better prediction intervals, which is what the methods are often used for in practice. Moreover, the current approach to test prediction intervals directly has additional flaws. We show why it is fundamentally flawed to test a prediction or confidence interval on a single test set. At best, marginal coverage is measured, implicitly averaging out overconfident and underconfident predictions. A much more desirable property is pointwise coverage, requiring the correct coverage for each prediction. We demonstrate through practical examples that these effects can result in favoring a method, based on the predictive uncertainty, that has undesirable behaviour of the confidence or prediction intervals. Finally, we propose a simulation-based testing approach that addresses these problems while still allowing easy comparison between different methods.
\end{abstract}
\end{frontmatter}

\section{INTRODUCTION}
\noindent Neural networks are, among other things, currently being used in a wide range of regression tasks, covering many different areas. It has become increasingly clear that it is essential to have uncertainty estimates to come with the predictions \citep{gal2016uncertainty, pearce2020uncertaintythesis}. Uncertainty estimates can be used to make confidence intervals or predictions intervals at a given $100 \cdot(1-\alpha)\%$ confidence level. We define these intervals more precisely in Section \ref{Defining uncertainty}, but the intuition is as follows. The probability that the true function value falls inside a confidence interval (CI) should be $100\cdot (1-\alpha) \%$. A $100\cdot(1-\alpha)\%$  prediction interval (PI) is constructed such that the probability that an observation falls inside this interval is $100\cdot(1-\alpha)\%$. The two desirable characteristics of a PI or CI are that they cover the correct fraction of the data while being as small as possible \citep{khosravi2011comprehensive}. At the moment, a common approach to test a PI is to use a previously unused part of the data and then check which fraction of the observations falls inside the corresponding PIs. This fraction is called the Prediction Interval Coverage Probability (PICP) and is widely used to asses the quality of prediction intervals \citep{pearce2018high, kabir2023Uncertainty, khosravi2011comprehensive, pearce2020uncertainty, su2018tight, lai2022exploring, zhang2023Probabilistic, chen2023Prediction, dewolf2023Valid, vanbeers2023Peaking, zhang2023Interval, zheng2023Stochastic} . 

While some methods, such as quantile regression, output PIs directly, others output a density. The testing procedure for these methods is largely influenced by the article of \citet{hernandez2015probabilistic}. To compare their proposed method, probabilistic backpropagation, with existing alternatives, they came up with a novel testing procedure. Their testing procedure uses ten publicly available real-world data sets and evaluates the loglikelihood on an unseen test set using a fixed training procedure. We explain this procedure in detail in Section \ref{logprecedure}. This setup allowed for an effective way of comparing different methods. Many authors subsequently used this setup as a benchmark to compare their uncertainty estimation methods with those of others \citep{gal2016dropout, lakshminarayanan2017simple, mancini2020prediction,liu2016stein,salimbeni2017doubly}. Recent work on PIs also uses these data sets to calculate the PICP score, see for example \citet{pearce2018high}. When using real-world data sets, it is impossible to directly evaluate the CI since the true function that generated the data is not known.
 
In this article, we demonstrate, both theoretically and through simulation experiments, that both testing methodologies fail to accurately determine the quality of a prediction or confidence interval. Specifically, we show that a better loglikelihood does not guarantee better PIs or CIs. Furthermore, covering the correct fraction of points in a test set does not test coverage correctly and even if it did, it does not guarantee that the PIs are correct for individual data points. As a result, it is possible to select the wrong method, that may not produce sensible PIs or CIs for individual points, as the best. 

This article consists of six sections, this introduction being the first. Section \ref{Defining uncertainty} gives the theoretical framework that is needed to properly discuss uncertainty. We precisely define what we mean with coverage and explain the loglikelihood and PICP testing approach. Section \ref{Shortcomings} gives theoretical drawbacks of these current testing methodologies. In Section \ref{Simulation}, we propose a simulation-based approach to combat these drawbacks. Section \ref{Experimental} verifies these concerns using simulation results by comparing the PICP approach to a simulation-based one. Finally, in Section \ref{Conclusion}, we summarise the conclusions and give suggestions for future work. 
\section{Defining the Uncertainty Framework and Testing Methodology}

\label{Defining uncertainty}
\noindent This section consists of three parts. First we go through the terminology necessary to properly discuss uncertainty. In the second and third part we explain the two most popular testing procedures, evaluating the loglikelihood and evaluating PICP. For an overview of the various methods to obtain uncertainty estimates, we refer to \citet{khosravi2011comprehensive} for early work, and to the reviews by \citet{he2023Survey, hullermeier2021aleatoric, gawlikowski2021survey,  kabir2018neural} for more recent contributions. 

\subsection{Defining Uncertainty}
Before we can talk about quantifying uncertainty, we need to define precisely what we mean with the term. Throughout this article we assume a regression setting. We have a data set $\mathcal{D} = \left((\bm{x}_{1}, y_{1}), \ldots (\bm{x}_{N}, y_{N})\right)$ that is a set of $N$ independent realisations of the random variable pair $(X,Y)$. We assume that $\bm{x} \in \mathbb{R}^{d}$ and $y\in \mathbb{R}$. The regression situation we are considering is such that
\[
Y \mid X=\bm{x} \sim \N{f(\bm{x})}{\sigma^{2}(\bm{x})},
\]
where $f(\bm{x})$ is the true regression function and $\sigma^{2}(\bm{x})$ is the variance of the additive noise. This is equivalent to the typical description of a regression setting where
we assume that our observations are a combination of an unknown function and a (normally distributed) noise term:
\[
y_{i} = f(\bm{x}_{i}) + \epsilon_{i}.
\]

Suppose we train a neural network (or any other type of model) to approximate $f(\bm{x})$ with $\hat{f}(\bm{x})$. We have two types of uncertainty. In the first place, we are unsure about the quality of our estimate $\hat{f}(\bm{x})$. We refer to this as the \textit{model} uncertainty (in other works sometimes referred to as \textit{epistemic} uncertainty). On the other hand, if we want to predict a new $y$-value, we have an additional source of uncertainty due to the inherent randomness of $\epsilon_{i}$. If the distribution of $\epsilon_{i}$ given $\bm{x}_{i}$ does not depend on $\bm{x}_{i}$, we have \textit{homoscedastic} noise. If the distribution of $\epsilon_{i}$ given $\bm{x}_{i}$ depends on $\bm{x}_{i}$, we have \textit{heteroscedastic} noise. The uncertainty due to $\epsilon_{i}$ is often referred to as the \textit{aleatoric uncertainty}, \textit{irreducible variance}, \textit{data noise variance}. In this paper, we use the terminology \textit{data noise variance}.

For an application of a model in practice, it is often necessary to quantify this uncertainty. Ultimately, one may want to do this by giving an accompanying confidence or prediction interval. We now take some time to define these concepts more precisely as this will be the cornerstone of the discussion in the rest of this paper.

We take a fixed covariates viewpoint. With $\mathscr{Y}$, we denote the random variables whose realisation is an entirely new set of targets $(y_{1}, \ldots, y_{n})$. This is equivalent to realisations of $Y \mid X={\bm{x}_{i}}$. We add this extra notation to distinguish taking an expectation over an entire new set of targets and taking an expectation over a single observation pair $(X,Y)$. For a given set of covariates, we have an estimator that is a function of the targets. In this case we may want to take the expectation over $\mathscr{Y}$. In a machine learning context, the predictor $\hat{f}$ often does not only depend on the data set but also on random effects, such as the weight initialisation of a neural network and the ordering in which the training examples are presented. With $U$, we denote a random variable that expresses randomness in a training process. We are now ready to define a confidence and prediction interval.\\

\begin{definition}
\label{CIpointwisedefinition}
	A $(1-\alpha) \cdot 100\%$ \textit{pointwise} confidence interval for $f$ is a random mapping, \\$\textnormal{CI}^{(\alpha)}(\mathscr{Y}, U, \cdot): \mathbb{R}^{d} \rightarrow  \mathcal{P}(\mathbb{R}) : \bm{x} \mapsto \textnormal{CI}^{(\alpha)}(\mathscr{Y}, U, \bm{x})$, such that
\begin{equation}
\label{CIpointeq}
	 \mathbb{E}_{\mathscr{Y},U} \left[ \mathds{1}_{\{f(\bm{x}) \in \textnormal{CI}^{(\alpha)}(\mathscr{Y}, U, \bm{x})\}} \right] = 1-\alpha \quad \forall  \bm{x}.
\end{equation}
\end{definition}
Here, $\mathcal{P}(\mathbb{R})$ is the power set of $\mathbb{R}$.  With $(\mathscr{Y},U)$, we explicitly denote that the construction of the interval depends on the specific realisation of the targets and of a random effect. For each realisation of these random variables, we get a different confidence interval. We drop this extra notation later on and simply write $CI^{(\alpha)}(\bm{x})$. Intuitively, this says that, given our set of covariates, if we randomly sample the targets and create a confidence interval, that the probability that $f(\bm{x})$ falls inside that interval is $1-\alpha$ for all values of $\bm{x}$. We refer to this type of coverage as $\textit{pointwise coverage}$. We also define $\textit{marginal coverage}$. It is less desirable, but implicitly often used (see Section \ref{Shortcomings}). \\
\begin{definition}
\label{CImarginaldefinition}
	A $(1-\alpha) \cdot 100\%$ marginal confidence interval for $f$ is a random mapping, \\$\textnormal{CI}^{(\alpha)}(\mathscr{Y}, U, \cdot): \mathbb{R}^{d} \rightarrow  \mathcal{P}(\mathbb{R}) : \bm{x} \mapsto \textnormal{CI}^{(\alpha)}(\mathscr{Y}, U, \bm{x})$, such that
\begin{equation}
\label{CIsimuleq}
	 \mathbb{E}_{\mathscr{Y},U} \left[\E_{{X}} \left[\mathds{1}_{\{f(X) \in \textnormal{CI}^{(\alpha)}(\mathscr{Y},U, X)\}} \right]\right] = 1-\alpha.
\end{equation}
\end{definition}

Marginal coverage states that the probability that, for a random realisation of $X$ and a random realisation of the confidence interval (which is random because the specific training set and training process is random), the function value falls inside the CI is $(1-\alpha) \cdot 100 \%$. The inner expectation gives the probability that a function value for a random realisation $\bm{x}$ falls in \textit{that} specific confidence interval. The outer expectation averages over all possible targets and random training effects, each resulting in slightly different intervals. We emphasize that in this definition the coverage may be very different across different values of $\bm{x}$. It is immediately clear from the definitions that pointwise coverage is much stronger than marginal coverage\footnote{An even stronger notion of coverage would be \textit{simultaneous} coverage, where the function values must fall in the corresponding intervals for all $x$-values \textit{at the same time} with probability $1-\alpha$. There is very little work on simultaneous intervals within the machine learning community so we do not elaborate on it further. We refer to \citet{degras2017simultaneous} for an example in a related field.}

Analogously we define a pointwise and marginal prediction interval as follows.
\\
\begin{definition} \label{PIpointwisedefinition}
	A $(1-\alpha) \cdot 100\%$ pointwise prediction interval is a random mapping,\\ $\textnormal{PI}^{(\alpha)}(\mathscr{Y}, U, \cdot): \mathbb{R}^{d} \rightarrow \mathcal{P}(\mathbb{R}) : \bm{x} \mapsto \textnormal{PI}^{(\alpha)}(\mathscr{Y}, U, \bm{x})$, such that	
	\begin{equation}
		\label{PIdef}
		\mathbb{E}_{\mathscr{Y},U} \left[ \mathbb{E}_{Y \mid X=\bm{x}}\left[ \mathds{1}_{\{Y \in \textnormal{PI}^{(\alpha)}(\mathscr{Y}, U, \bm{x})\}}\right] \right]= 1-\alpha \quad \forall \bm{x}.
	\end{equation}
\end{definition}

\begin{definition} \label{PIdefinition}
	A $(1-\alpha) \cdot 100\%$ marginal prediction interval is a random mapping,\\ $\textnormal{PI}^{(\alpha)}(\mathscr{Y}, U, \cdot): \mathbb{R}^{d} \rightarrow \mathcal{P}(\mathbb{R}) : \bm{x} \mapsto \textnormal{PI}^{(\alpha)}(\mathscr{Y}, U, \bm{x})$, such that	
	\begin{equation}
	  \mathbb{E}_{\mathscr{Y},U} \left[\E_{{X,Y}} \left[\mathds{1}_{\{Y \in \textnormal{PI}^{(\alpha)}(\mathscr{Y},U, X)\}}\right] \right] = 1-\alpha.
	\end{equation}
\end{definition}

The difference between the two can be exemplified with a weather forecast. Suppose the weatherman gives a 90$\%$ prediction interval for the temperature tomorrow. If this is a pointwise prediction interval, then the probability that the true temperature tomorrow falls inside that interval is 90$\%$. If it is a marginal interval, however, then the weatherman says that averaged over all possible days, the temperature will fall in those intervals in 90$\%$ of the time, but there is no real guarantee for tomorrow. In the second case, the weatherman is allowed to be 85$\%$ correct in winter and $95\%$ in the summer. The weatherman could even simply pick 3 days each month to give a point estimate - and thus being always wrong - and give the interval from -100 to 100 degrees celsius for the rest of the month. Since these intervals are not very useful, we favor smaller intervals. We note that, for some applications, marginal coverage may be adequate and that it sometimes is explicitly the goal (e.g., with split-conformal inferencer).

In the following two subsections, we examine the most popular testing procedures for uncertainty estimates. We identify two different strategies, which we will explain in order:

\begin{enumerate}

\item The method outputs a prediction interval and the relevant metrics are the average width of the intervals and the fraction of test points that fall inside the prediction intervals (PICP).
\item The method outputs a density $p(y \mid \bm{x})$, generally $\N{\hat{f}(\bm{x})}{\hat{\sigma}^{2}(\bm{x})}$ and the relevant metrics are the loglikelihood of a test set and the root mean squared error, RMSE.
\end{enumerate}

\subsection{The PICP Testing Procedure}
We first explain a popular method to test prediction intervals directly. The idea is to take a real-world data set, split it in a training and test set, create prediction intervals using the training set, and calculate the fraction of test points that falls inside the prediction intervals. The relevant metric in this case is the PICP.
\begin{definition}
The Prediction Interval Coverage Probability, or \textnormal{PICP}, is the fraction of observations in a test set that falls inside the corresponding prediction intervals:
\[
	\textnormal{PICP}  :=\frac{1}{N_{\textnormal{test}}} \sum_{i=1}^{N_{\textnormal{test}}} \mathds{1}_{\{y_{i} \in \textnormal{PI}(\bm{x}_{i})\}}
\]
\end{definition}
Note that every method can be tested this way. Methods that output a density can create a PI using that density. This PI can be compared with the PI of a method that directly outputs one (such as quantile regression for instance). We can define the same measure for a confidence interval.\\
\begin{definition}
The Confidence Interval Coverage Probability is the fraction of function values that falls inside the corresponding confidence intervals.
\[
	\textnormal{CICP}  :=\frac{1}{N_{\textnormal{test}}} \sum_{i=1}^{N_{\textnormal{test}}} \mathds{1}_{\{f(x_{i}) \in \textnormal{CI}(\bm{x}_{i})\}}.
\]
\end{definition}
 To be able to compute the CICP, one needs to have access to the true function $f(\bm{x})$. The average width of the prediction intervals is usually also reported since we prefer intervals that capture the correct fraction of the data while being as narrow as possible.

\subsection{The Loglikelihood Testing Procedure} \label{logprecedure}
The loglikelihood testing approach assumes the uncertainty estimation method outputs a density $p(y \mid \bm{x})$, usually a normal distribution with mean $\hat{f}(\bm{x})$ and standard deviation $\hat{\sigma}_{\text{predictive}}$. The objective is to get the highest average loglikelihood on the test set:
\[
\resizebox{0.48\textwidth}{!}{$
\text{LL} = \frac{1}{N_{\text{test}}} \sum_{i=1}^{N_{\text{test}}} \log\left[\frac{1}{\sqrt{2\pi \sigma^{2}_{\text{predictive}}}} \exp\left( -\frac{1}{2}\left(\frac{y_{i} - \hat{f}(\bm{x}_{i})}{\sigma_{\text{predictive}}} \right)^{2}\right) \right].$}
\]
Evaluating the likelihood tests how well the predicted density matches the true data generating density. This density could subsequently be used to make prediction intervals.

This approach has been popularised by \citet{hernandez2015probabilistic}. In their paper, they tested their method, probabilistic backpropagation, on ten publicly available real-world data sets (see Table \ref{Tab: datasets}). Using these data sets, they carried out the following procedure.
\begin{description}
    \item[Step 1:] Standardize the data so that it has zero mean and unit variance.
    \item[Step 2:] Split the data in a training and test set. They apply a 90/10 split.
    \item[Step 3:] Train the network using 40 epochs and update the weights of the network after each data point. They use a network with one hidden layer containing 50 hidden units. For the two largest data sets, \textit{Protein Structure} and \textit{Year Prediction MSD}, they chose 100 hidden units. 
    \item[Step 4:] Repeat steps 2 and 3 a total of 20 times and report the average root mean squared error (\textit{RMSE}) and loglikelihood (\textit{LL}) on the test set and their standard deviations. They undo the standardization before calculating the \textit{RMSE} and \textit{LL}.    
\end{description}
\begin{table*}[tb]
\centering	
\begin{tabular}{ l r r p{0.57\linewidth}} 
\textbf{Data set} & $N$ & $d$ & Description \\
 \hline
\textit{Boston Housing}  &506 &13 & Housing prices in suburbs of Boston as a function of covariates such as crime rates and mean number of rooms\\ 
\textit{Concrete Compression Strength}  &1030 &8 & Concrete compressive strength as a function of covariates such as temperature and age\\
\textit{Energy Efficiency}  &768 &8 & The energy efficiencies of buildings as a function of covariates such as wall area, roof area, and height\\ 
\textit{Kin8nm}  &8192 &8 & The forward kinematics of an 8 link robot arm \\ 
\textit{Naval propulsion}  &11,934 &16 & A simulated data set giving the propulsion behaviour of a naval vessel \\
\textit{Combined Cycle Power Plant}  &9568 &4 & The net hourly electrical energy output as a function of temperature, ambient pressure, relative humidity, and exhaust vacuum\\
\textit{Protein Structure}  &45,730 &9 & Physicochemical properties of protein tertiary structure \\ 
\textit{Wine Quality Red}  &1599 &11 & Wine quality as a function of physicochemical tests such as density, pH, and sulphate levels\\  
\textit{Yacht Hydrodynamics}  &308 &6 & Air resistance of sailing yachts as a function of covariates such as length-beam ratio, prismatic coefficient, or beam-draught ratio\\ 
\textit{Year Prediction MSD} & 515,345& 90 & The release year of a song based on audio features \\
\hline
\end{tabular}
\caption{The ten different regression data sets that are currently being used as a benchmark for estimating the quality of uncertainty estimates. The number of instances, $N$, number of covariates, $d$, and a short description are given. }
\label{Tab: datasets}
\end{table*}

This setup has subsequently been used in multiple articles \citep{gal2016dropout,lakshminarayanan2017simple,mancini2020prediction,liu2016stein,salimbeni2017doubly}.  Recently, these data sets have also been used to calculate the PICP metric \citep{khosravi2011comprehensive, pearce2020uncertainty, su2018tight, lai2022exploring}. The following section discusses shortcomings of both the PICP and loglikelihood approach.

\section{THEORETICAL SHORTCOMINGS OF THE CURRENT TESTING METHODOLOGY} \label{Shortcomings}
\noindent In this section, we discuss four problems with the aforementioned methods of testing the quality of uncertainty estimates on a single test set with the PICP or the loglikelihood. 

\subsection{Predictive Performance Does Not Guarantee Good Model Uncertainty Estimates} \label{whichestimate}
For some applications, the predictive uncertainty is the only relevant quantity. For the prices on the stock market, it does not matter what the underlying function was, the actual observation is what counts. For a physicist trying to measure a constant or functional relation, however, the model uncertainty may be much more relevant. This model uncertainty is often used for out-of-distribution detection. The reasoning is that in an area that is previously unseen by the model, the model uncertainty is likely to be high. It is therefore crucial to know if the model uncertainty estimate is correct or not.

It is implicitly assumed that methods that have a better predictive performance on a test set also estimate the model uncertainty better. This need not be the case. Suppose we have separate estimates for the data noise variance and model uncertainty. These two estimates can be combined to obtain a predictive uncertainty estimate. Now we compare two methods, $A$ and $B$,  by carrying out the tests as described in the previous section, either the PICP or the loglikelihood. If method $A$ gets a better score than method $B$, it is unclear why. It is possible that the estimate for the model uncertainty in method $A$ was much worse than for method $B$, but that this was compensated by a superior estimate of the data noise variance. Yet another possibility is that both estimates are incorrect but result, more or less, in the correct total uncertainty. We provide empirical support in Section \ref{Experimental}. 
\subsection{Coverage Cannot Be Tested on a Single Data Set} \label{singleset}
A confidence or prediction interval is a random variable because it depends on the specific realisation of the targets. In the context of a neural network, there is also an added random training aspect. It is therefore necessary to test these intervals by repeating the entire process multiple times. This means that new targets need to be collected, a new network needs te be trained, and a new interval needs to be constructed. For a PI, for instance, the marginal coverage can be approximated by evaluating the PICP multiple times:
\[
\frac{1}{L}\sum_{l=1}^{L} \left( \frac{1}{N_{\text{test}}} \sum_{i=1}^{N_{\text{test}}} \mathds{1}_{\{y_{i} \in \textnormal{PI}_{l}(\bm{x}_{i})\}} \right) = \frac{1}{L}\sum_{l=1}^{L} \left( \text{PICP}_{l} \right),
\]
where $L$ is the total number of repeated experiments and the subscript $l$ indicates that the prediction intervals will be different in each simulation. In fact, if we do this infinitely many times with an infinitely large test set, this approximation becomes exact since by the law of large numbers

\begin{equation*}
\begin{gathered}
\lim_{L \rightarrow \infty} \lim_{N_{\text{test}} \rightarrow \infty} \frac{1}{L}\sum_{l=1}^{L} \left( \frac{1}{N_{\text{test}}} \sum_{i=1}^{N_{\text{test}}} \mathds{1}_{\{y_{i} \in \textnormal{PI}_{l}(\bm{x}_{i})\}} \right) \\
= \mathbb{E}_{\mathscr{Y},U} \left[\E_{{X, Y}}\left[ \mathds{1}_{\{Y \in \textnormal{PI}^{(\alpha)}(\mathscr{Y},U, X)\}} \right] \right].
\end{gathered}
\end{equation*}
By examining the previous equation, we observe that the PICP in fact only gives a single approximation of the inner expectation term, $\E_{{X, Y}} \left[\mathds{1}_{\{Y \in \textnormal{PI}^{(\alpha)}(\mathscr{Y},U, X)\}}\right]$. 

\begin{figure}[tb]
	\centering
	\includegraphics[width = 8.5cm]{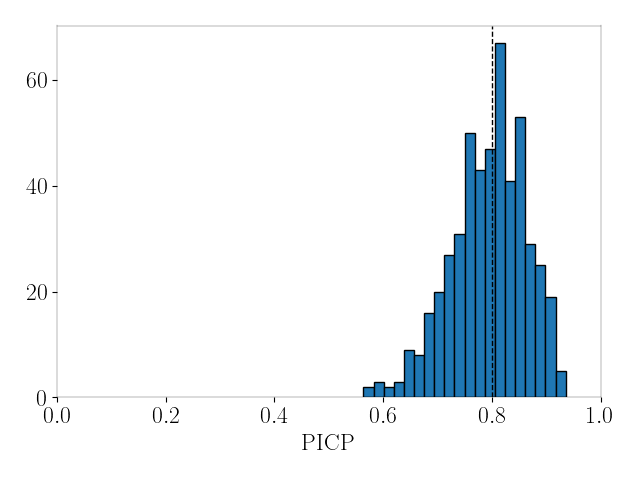}
	\caption{The PICP values of 500 simulations. In each simulation, new data was generated, a new linear model was fit, and a new prediction interval was created.}
	\label{linearmodelPICP}
\end{figure}

To exemplify this problem, we simulated an example where we fitted a linear model on 25 data points. The $x$-values were simulated uniformly between -2 and 2 and  we used $Y \mid X = x \sim \N{x}{0.1^{2}}$, a straight line with some noise. With these 25 data points, we constructed an 80$\%$ prediction interval using classical theory and computed the PICP using a test set containing 500 data points. We repeated this process 500 times to demonstrate that a single evaluation of the PICP is not indicative of the quality of a prediction interval. The PICP values are shown in Figure \ref{linearmodelPICP}. We have a perfect prediction interval but the individual PICP values range between 0.58 and 0.92. 

Analogously, we see that the CICP gives a single realisation of the inner expectation in the definition of marginal coverage:
\begin{equation*}
\begin{gathered}
\lim_{L \rightarrow \infty} \lim_{N_{\text{test}} \rightarrow \infty} \frac{1}{L}\sum_{l=1}^{L} \left( \frac{1}{N_{\text{test}}} \sum_{i=1}^{N_{\text{test}}} \mathds{1}_{\{f(\bm{x}_{i}) \in \textnormal{CI}_{l}(\bm{x}_{i})\}} \right) \\= \mathbb{E}_{\mathscr{Y},U} \left[\E_{{X}} \left[\mathds{1}_{\{f(X) \in \textnormal{CI}^{(\alpha)}(\mathscr{Y},U, X)\}} \right]\right].
\end{gathered}
\end{equation*}
Looking at a linear model illustrates this can be even more problematic. Consider a linear model without a bias term,
\[
y_{i} = ax_{i} + \epsilon_{i}.
\]
The model uncertainty considers the uncertainty in our estimate $\hat{a}$. For this example, we assume that the true function is of the form $f(x) = ax$, meaning that the true function is within our hypothesis class. Suppose we have a perfectly calibrated procedure to construct a $95 \%$ CI for $a$. This means that an interval constructed by that procedure will contain the true value of $a$ in $95\%$ of the experiments (collecting data, fitting the model, creating the CI).

\begin{figure*}[h]
	\centering
	\includegraphics[width = 13cm]{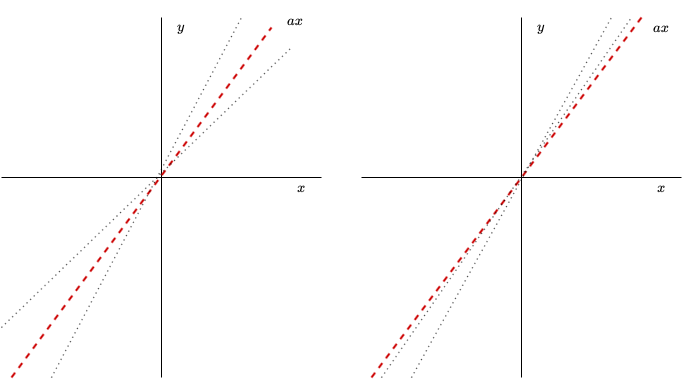}
	\caption{The dashed red line gives the true function $f(x) = ax$. The two dotted black lines give the confidence interval. In the left figure the true function falls inside our confidence interval, in the right figure it does not. It is clear that the measure CICP will either give 1 or 0, even when we have a method that gives a perfect $95\%$ confidence interval: It is impossible to test the coverage of our CI on a single test set.}
	\label{linearmodelexample}
\end{figure*}

We can translate the CI of $a$ to a CI of $f(x_{i}) = ax_{i}$. However, since the true $a$ is either inside our CI or not, $f(x_{i})$ is for all $x_{i}$ either inside the CI or not. This results in a CICP of either 0 or 1, even though the uncertainty estimate is perfectly calibrated. This illustrates that we cannot test the quality of our CI by simply looking at the fraction of points in a single test set that are in our interval. This example is illustrated in Figure \ref{linearmodelexample}. It is simply not possible to test the quality of a prediction or confidence interval on a single data set.

\subsection{A Good PICP Score Does Not Guarantee Pointwise Coverage}

The PICP score estimates marginal coverage and not pointwise coverage. It is desirable to have prediction and confidence intervals that have the correct coverage for each specific value of $\bm{x}$ and not merely on average. 

\begin{figure}[t]
	\centering
	\includegraphics[width = 8.5cm]{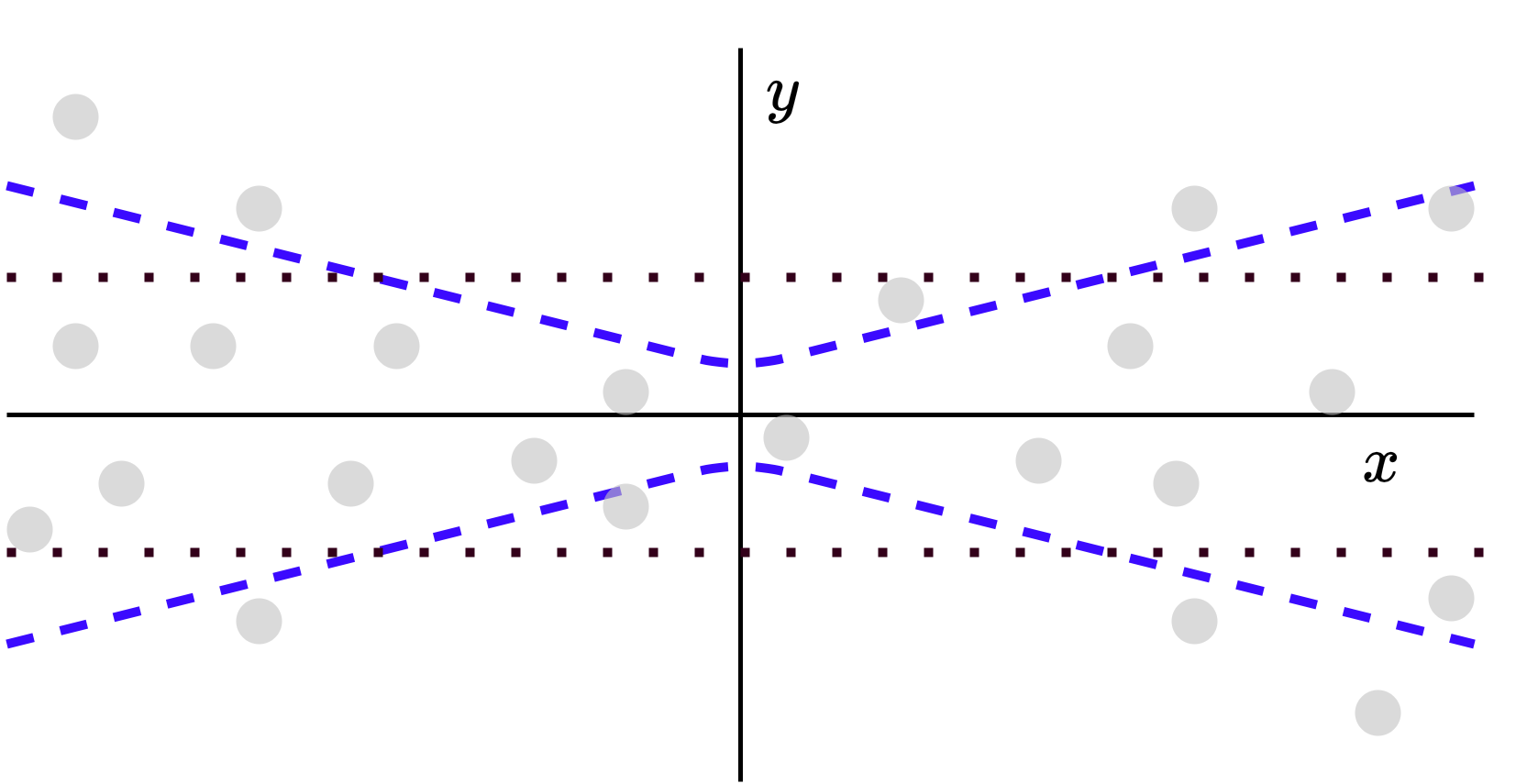}
	\caption{This figure illustrates that estimating the data noise variance correctly can result in a PICP close to the chosen confidence level. We assume that the model uncertainty is comparably very small. The true function, $f(x)$, is the constant zero function. The dashed blue line gives $\pm 1\sigma(x)$. The dotted black line gives $\pm 1 \hat{\sigma}(x)$. On average, the data noise variance estimate is correct and its corresponding PI captures the correct fraction of the data in this case. Using PICP in this example, we do not notice that our uncertainty estimate is wrong.}
	\label{linearmodelexample2}
\end{figure}

Marginal and pointwise coverage can be related as follows:
\begin{equation}
\begin{gathered}
\label{marginalpointwise}
	\int_{X,Y} \mathds{1}_{y \in \textnormal{PI}(\bm{x})} \pi(\bm{x},y)d\bm{x}dy \\ =\int_{X} \left( \int_{Y}\mathds{1}_{y \in \textnormal{PI}(\bm{x})}\pi(y \mid \bm{x})dy \right) \pi(\bm{x})d\bm{x}.
\end{gathered}
\end{equation}
This illustrates that a good marginal coverage does not imply a good pointwise coverage. It is possible to get a good PICP score (which approximates the left integral in Equation \eqref{marginalpointwise}) while only estimating the predictive uncertainty correctly \textit{on average}. We illustrate this in Figure \ref{linearmodelexample2}. Assume that the true function is the constant zero function, $f(x) = 0$, and that our estimate of the function is very good, $\hat{f}(x) \approx 0$. The dashed blue lines give plus and minus one time the true standard deviation of the data. Suppose that we use a homoscedastic estimate of this standard deviation, the dotted black line,  and that this estimate has on average the correct size. The PICP score of the $68\%$ PI made with our homoscedastic estimate is 0.65 in this example. The intervals are too wide for some $x$ and too small for others but \textit{on average} capture the correct fraction of the data. We are not able to see that our estimate is wrong by simply evaluating the PICP on a single test set.

We also want to stress that, with a one-dimensional output, it is often possible to get close to the desired PICP value on an unseen part of the training set by tuning the hyperparameters. Monte Carlo dropout for instance has the data noise variance as a hyperparameter. If the PICP value is too low, it is possible to tune this parameter until it is correct. If the test set resembles the training set, it is unsurprising that a good PICP score can also be achieved on this set. In Section \ref{Experimental} we elaborate on this. 

\subsection{A Better Loglikelihood Does Not Imply Better Prediction Intervals} \label{testt}

Before addressing the downside of the loglikelihood, we need to mention the argument in favor of using it. If the goal is to find the density that is closest to the true density $\pi(y\mid \bm{x})$, then the loglikelihood is optimal in the following sense. Suppose that the outputted density is parametrized by $\theta$ and we find $\hat{\theta}$ such that the loglikelihood is maximal on a test set: 
\[
\hat{\theta} = \arg\max_{\theta} \sum_{i=1}^{N_{\text{test}}} \log(p_{\theta}(
\bm{x}_{i}, y_{i})).
\]
\citet{akaike1973information} showed that (under some assumptions) $\hat{\theta}$ is a natural estimator for the $\theta$ that minimises the KL distance between the true density and the outputted density. In this sense, the loglikelihood seems to be the obvious metric to measure the quality of an uncertainty estimate. However, if the eventual goal is to make a prediction interval, which is often the case in applications, then the loglikelihood can easily favor the method that produces worse prediction intervals. This is because although loglikelihood depends on the quality of the predicted variance, it also highly depends on the quality of the fit. A higher loglikelihood can therefore be the result of a better fit or of a better estimate of the predictive variance.

This ambiguity can be problematic since the method with the higher loglikelihood can produce significantly worse prediction intervals. To show this, we go through a quick example where two methods are compared by using the loglikelihood on a test set. Suppose we have $N$ data points, $x_{i}$, from a $\N{100}{5^2}$ distribution. We want to compare two blackbox methods that output a mean estimate, $\hat{\mu}$, and an uncertainty estimate, $\hat{\sigma}^{2}$. The blackbox methods arrive at the following estimates:
\[
\hat{\mu}_{1} = \frac{1}{N} \sum_{i=1}^{N} x_{i}, \quad \hat{\sigma}_{1} = 0.9 \sqrt{\frac{1}{N} \sum_{i=1}^{N}(x_{i} - \hat{\mu}_{1})^{2}},
\]
and
\[
\hat{\mu}_{2} = 1.05 \frac{1}{N} \sum_{i=1}^{N} x_{i}, \quad \hat{\sigma}_{2} = \sqrt{\frac{1}{N} \sum_{i=1}^{N}(x_{i} - \hat{\mu}_{2})^{2}}.
\]
The mean estimate of model 1 is better than that of model 2, but its uncertainty estimate is too small. With these estimates, we create $68\%$ prediction intervals for both models:
\[
\textnormal{PI}_{i} = \hat{\mu}_{i} \pm \hat{\sigma}_{i}.
\]
 We ran 10000 simulations to compute the coverage of the prediction intervals. This means simulating new data, getting new estimates, and creating a new PI. In this example, model 1 has a slightly better loglikelihood (-3.109 versus -3.110) but a considerably worse coverage (0.57 versus 0.67).

This demonstrates that it is important to clearly have in mind what the end goal of the uncertainty quantification method is. If the end goal is to find the density closest to the true density, then the loglikelihood is a useful metric. If, however, the eventual goal is to construct prediction intervals, the same loglikelhood can be misleading.

 A practical example can be found in the paper on Monte Carlo dropout \citep{gal2016dropout}. The authors compare their method to an, at that time, popular variational inference method \citep{graves2011practical} and probabilistic backpropagation \citep{hernandez2015probabilistic}. Monte Carlo dropout achieves superior or comparable loglikelihood scores on the test sets of all ten data sets in Table \ref{Tab: datasets}. However, MC dropout also obtains the lowest RMSE on the test sets for nine of the ten data sets. It is not immediately clear if the better loglikelihood is the result of more precise predictions or of a better uncertainty estimate. Subsequently, the paper on Deep Ensembles \citep{lakshminarayanan2017simple} outperformed MC dropout in terms of both loglikelihood and RMSE, leaving the question open if the uncertainty estimate is actually better.

We therefore argue that if the eventual goal is to make prediction or confidence intervals, the best testing approach is to test \textbf{pointwise coverage}. A good pointwise coverage ensures good marginal coverage as well. In the following section, we explain an approach that allows us to test pointwise coverage: simulation-based testing. 

\section{SIMULATION-BASED TESTING} \label{Simulation}

\noindent In this section, we propose a new testing approach that addresses the issues that were raised in the previous section. More specifically, we give a testing procedure that tests pointwise coverage in a correct manner. As we argued in the previous section, it is impossible to test coverage correctly on a single data set. It is necessary to repeat the entire experiment multiple times, with new data sets, and then measure the pointwise coverage. We therefore propose a simulation-based setup, in which we are able to simulate new data sets and know the true data generating distribution. This has two advantages:
\begin{enumerate}
	\item The experiment can be repeated multiple times. This allows us to test coverage in the correct sense.
	\item The true function, $f(\bm{x})$, is known. This allows us to directly test the quality of a confidence interval, and not only a prediction interval.
\end{enumerate}
The metrics we propose are the Prediction Interval Coverage Fraction (PICF), the Confidence Interval Coverage Fraction (CICF), and the average width of the intervals. The PICF is defined as follows.
\begin{equation}
\label{PICFequation}
\text{PICF}(\bm{x}) := \frac{1}{N_{\text{sim}}} \sum_{s = 1}^{N_{\text{sim}}}\E_{Y \mid X=\bm{x}} \left[\mathds{1}_{\{Y \in \textnormal{PI}_{s}(\bm{x})\}}\right],
\end{equation}
where $Y$ is the random variable of which the realisations are the observations, $N_{\text{sim}}$ the number of simulations, and $\textnormal{PI}_{s}(\bm{x})$ the prediction interval for $\bm{x}$ in simulation $s$. Our proposed approach effectively gives a Monte Carlo approximation of the pointwise coverage. For a large number of simulations this precisely becomes the definition of pointwise coverage by the law of large numbers:
\begin{equation*}
\begin{gathered}
\lim_{N_{\text{sim}} \rightarrow \infty} \frac{1}{N_{\text{sim}}} \sum_{s = 1}^{N_{\text{sim}}}\E_{Y \mid X=\bm{x}} \left[ \mathds{1}_{\{Y \in \textnormal{PI}_{s}(\bm{x})\}}\right]\\ = \mathbb{E}_{\mathscr{Y},U} \left[ \E_{Y \mid X=\bm{x}}\left[\mathds{1}_{\{Y \in \textnormal{PI}(\mathscr{Y},U, \bm{x})\}}\right] \right]. 
\end{gathered}
\end{equation*}
This setup forces us to manually define the data generating process. If we use
\[
\mathcal{P}_{Y \mid X =\bm{x}} = \N{f(\bm{x})}{\sigma^{2}(\bm{x})},
\] 
then the expectation in Equation \eqref{PICFequation} can be calculated as
\begin{equation}
\begin{gathered}
\E_{Y \mid X=\bm{x}}\left[ \mathds{1}_{\{Y \in \textnormal{PI}(\bm{x})\}}\right] \\= \Phi\left(\frac{R^{(s)}(\bm{x}) - f(\bm{x})}{\sigma(\bm{x})}\right) - \Phi\left(\frac{L^{(s)}(\bm{x}) - f(\bm{x})}{\sigma(\bm{x})}\right),
\end{gathered}
\end{equation}
where $\Phi$ is the CDF of a standard normal Gaussian and $L^{(s)}(\bm{x}), R^{(s)}(\bm{x})$ are the lower and upper bounds respectively of the PI in simulation $s$. The superscript $s$ indicates that these intervals are different for each simulation. Analogously, we define the Confidence Interval Coverage Fraction:
\[
\text{CICF}(\bm{x}) := \frac{1}{N_{\text{sim}}} \sum_{s = 1}^{N_{\text{sim}}}\mathds{1}_{\{f(\bm{x}) \in \textnormal{CI}_{s}(\bm{x}) \}},
\]
where $f(\bm{x})$ is the true function value, and $LC^{(l)}(\bm{x}), RC^{(l)}(\bm{x})$ are the lower and upper limit of the CI of $f(\bm{x})$.  Note that if we have $N_{\text{test}}$ observations in our test set, then we obtain $N_{\text{test}}$ different computations of PICF($\bm{x}$) and CICF($\bm{x}$). These evaluations can be plotted in a histogram to see if the coverage fractions match the chosen confidence levels. In a one-dimensional setting, we can also plot the PICF and CICF as a function of $x$. 

If we construct a $100\cdot (1-\alpha)\%$ PI or CI, then we would ideally want the PICF($\bm{x}$) and the CICF($\bm{x}$) to be $1-\alpha$ for every $\bm{x}$. As a quantitive measure for the quality of the PI and CI we propose to use the Brier score, where lower is better. For the PICF, this yields 
\begin{equation}
\text{BS} = \frac{1}{N_{\text{test}}}\sum_{i = 1}^{N_{\text{test}}}\left(\text{PICF}(\bm{x}_{i}) -(1-\alpha)\right)^{2}.
\label{BS}	
\end{equation}
A Brier score for the PICF for instance is lower if the average is close to the desired value of $(1-\alpha)$ while having a low variance. We observe this by looking at the bias-variance decomposition
\begin{dmath*}
\E_{X}\left[\left(\text{PICF}(\bm{x}) - (1-\alpha)\right)^{2}\right] = \E_{X}\left[\text{PICF}(\bm{x}) - (1-\alpha)\right]^{2} + \mathbb{V}_{X}\left[\text{PICF}(\bm{x})\right].
\end{dmath*}
 This means that simply being correct on average instead of for all $\bm{x}$ results in a worse score when using the Brier score of the PICF. We can use the same measure to quantify the quality of a CI. The suggested simulation-based approach can be summarized as follows. \\

\begin{description}
	\item[Step 1:] Choose a distribution $\mathcal{P}_{X, Y}$ to simulate data sets from.
    \item[Step 2:] Simulate a test set. This test set does not change and must be the same when comparing different methods.
     \item[Step 3:] Simulate a training set and use the uncertainty estimation method to obtain the PIs and CIs at different confidence levels, for instance  95, 90, 80, and 70$\%$. Repeat this 100 times\footnote{Of course, more is better but we found that this works well enough for a comparison.}.
    \item[Step 4:] Calculate the PICF and CICF for each $\bm{x}$-value in the test set.
    \item[Step 5:] Evaluate the relevant metrics, for instance the Brier score and the average width.    
\end{description}
To simulate data, we propose three possibilities. The first option is to take a known test function and a noise term. The advantage of this is having total control over the setup. The disadvantage is that it may not be representative of real-world situations. A second option is to use simulations that are based on real-world data sets. The current benchmark data sets are good candidates. In the last part of the next section, we demonstrate a simulation setup for the popular Boston Housing data set. A third option is to use an extremely large data set. This data set can be split in 1 test set and 100 distinct training sets. The previous procedure can now be repeated but instead of simulating data we can use the real data. The disadvantage is that only prediction intervals can be tested in this way since the true underlying function is unknown.

\section{EXPERIMENTAL DEMONSTRATION OF THE ADVANTAGES OF SIMULATION-BASED TESTING} \label{Experimental}
\noindent In this section, we experimentally demonstrate the theoretical shortcomings raised in Section \ref{Shortcomings} and show how these issues can be resolved by using a simulation-based approach. As an illustration, we use two methods that are easy to implement: the na\"ive bootstrap \citep{heskes1997practical}, and concrete dropout \citep{gal2017concrete}, an improvement of the popular Monte Carlo dropout method \citep{gal2016dropout}. 

With these two approaches, we first show that a good PICP does not imply that the individual estimates for the data noise variance and model uncertainty are correct. Since we know the true underlying function, we can look at the CICP directly and verify its correctness. Secondly, we demonstrate the advantage of our proposed procedure to average over simulations per $\bm{x}$-value, instead of averaging over $\bm{x}$-values in a single test set. We demonstrate that this is useful by observing that having the desired CICP gives no guarantees that the CIs are correct for an individual $\bm{x}$-value. Most importantly, we demonstrate that it is possible to favor the wrong method when using the predictive performance (such as the PICP or loglikelihood) on a test set as the measure. We end this section by setting up a simulation based on the Boston Housing data set. 

\subsection{Concrete Dropout and the Naive Bootstrap} \label{dropout and bootstrap}
The uncertainty estimation methods used in this section assume that both the data noise variance and model uncertainty are normally distributed:
\[
y_{i} = f(\bm{x}_{i}) + \epsilon_{i}, \quad \text{with } \epsilon_{i} \sim \N{0}{\sigma^{2}(\bm{x}_{i})}
\]
and 
\[
\hat{f}(\bm{x}) = f(\bm{x}) + \epsilon_{\omega, i}, \quad \text{with }\epsilon_{\omega, i}\sim \N{0}{\sigma^{2}_{\omega}(\bm{x}_{i})}. 
\]
These two uncertainties can be combined to  jointly make up the total predictive uncertainty. The methods additionally assume that both uncertainty estimates are independent. This independence allows us to add up both the variances to obtain the variance of the predictive uncertainty. 

We use the bootstrap setup from \citet{heskes1997practical}. This setup outputs a CI as described in Algorithm \ref{Naivebootstrap}.
\\
\\
\begin{algorithm}[tb]
  \For{i \text{in} 1:$M$}{
  Resample $(X,Y)$ pairwise with replacement, denote this sample with $(X^{(i)}, Y^{(i)})$\;
  Train an ANN on $(X^{(i)}, Y^{(i)})$ that outputs $\hat{f}_{i}(\bm{x})$\;
  }
  Define $\hat{f}(\bm{x}) = \frac{1}{M} \sum_{i = 1}^{M} \hat{f}_{i}(\bm{x})$\;
  Calculate $\hat{\sigma}_{\omega}^{2}(\bm{x}) = \frac{1}{M - 1} \sum_{i = 1}^{M} \left(\hat{f}_{i}(\bm{x}) - \hat{f}(\bm{x}) \right)^{2}$\;
  CI$(\bm{x})$ = $[\hat{f}(\bm{x}) - t_{1- \alpha /2}^{M}\hat{\sigma}_{\omega}(\bm{x}),\;\hat{f}(\bm{x}) + t_{1- \alpha /2 }^{M}\hat{\sigma}_{\omega}(\bm{x})]$\;
  \textbf{return:} CI$(\bm{x})$\;
   \caption{Pseudocode to obtain a CI for $f(\bm{x})$ using an implementation of the na\"ive bootstrap approach as described by \citet{heskes1997practical}.}
  \label{Naivebootstrap}
\end{algorithm}
\

In this algorithm, $t_{1- \alpha /2 }^{M}$ is the $1-\frac{\alpha}{2}$ quantile of a $t$-distribution with $M$ degrees of freedom. We note that the variance in line 5 technically gives the variance of an individual ensemble member and not of the average. As we will see in Section 5, this results in confidence intervals that are often too large.

We train networks with $40, \;30, $ and $ 20$ neurons respectively and $ReLU$ activation functions for $80$ epochs. In order to arrive at a prediction interval we also need an estimate of the data noise variance. Assuming homoscedastic noise, we take 
\begin{equation}
\hat{\sigma}^{2} = \frac{1}{N_{\text{val}}}\sum_{j = 1}^{N_{\text{val}}} \max\left(\left( y_{j} - \hat{f}(\bm{x}_{j})\right)^{2} -  \hat{\sigma}_{\omega}^{2}(\bm{x}) , 0 \right).
\label{aleatoricestimate}
\end{equation}
Note that we use a small additional validation set to determine $\hat{\sigma}$. With both $\hat{\sigma}$ and $\hat{\sigma}_{\omega}(\bm{x})$, we construct the prediction interval
\[
\resizebox{0.48\textwidth}{!}{$\textnormal{PI}(\bm{x}) = \Big[\hat{f}(\bm{x}) - t_{1- \alpha /2}^{M}\sqrt{\hat{\sigma}_{\omega}^{2}(\bm{x}) + \hat{\sigma}^{2}} , \;\hat{f}(\bm{x}) + t_{1- \alpha /2 }^{M}\sqrt{\hat{\sigma}_{\omega}^{2}(\bm{x}) + \hat{\sigma}^{2}}\Big]$}.
\]

A different approach to obtain uncertainty estimates is Monte Carlo dropout \citep{gal2016dropout}. The easy implementation makes this method very popular. The idea is to train a network with dropout enabled and then keep dropout enabled while making predictions. The article shows that, under certain conditions, this is equivalent to sampling from an approximate posterior. The standard deviation of these forward passes is used as an estimate of the model uncertainty. The inverse of the hyperparameter $\tau$ gives the estimate of the variance of the noise. A follow-up paper \citep{gal2017concrete} further refines the method. This so called \textit{concrete} dropout does not rely on a hyperparameter to estimate the data noise variance but outputs a heteroscedastic estimate directly. Additionally, the dropout probability is tuned as a part of the training process. Algorithm \ref{dropout} describes the procedure in more detail.

\begin{algorithm}[tb]
  Train an ANN with dropout enabled on $\mathcal{D}_{\text{train}}$. This network has two output neurons corresponding to $\hat{f}_{b}(\bm{x}_{i})$ and $\hat{\sigma}^{2}_{b}(\bm{x}_{i})$ and is trained by maximizing the loglikelihood assuming a normal distribution. The $b$ subscript indicates that each forward pass gives a different result.\;
  Define $\hat{f}(\bm{x}) = \frac{1}{B} \sum_{b = 1}^{B} \hat{f}_{b}(\bm{x})$\;
   Define $\hat{\sigma}^{2}(\bm{x}) = \frac{1}{B} \sum_{b = 1}^{B} \hat{\sigma}^{2}_{b}(\bm{x})$\;
  Define $\hat{\sigma}_{\omega}^{2}(\bm{x}) := \frac{1}{B - 1} \sum_{b = 1}^{B} \left(\hat{f}(\bm{x}) -  \hat{f}_{b}(\bm{x})\right)^{2}$\;
  CI$(\bm{x})$ = $[\hat{f}(\bm{x}) - z_{1- \alpha /2}\hat{\sigma}_{\omega}(\bm{x}),\;\hat{f}(\bm{x}) + z_{1- \alpha /2 }\hat{\sigma}_{\omega}(\bm{x})]$\;
 PI$(\bm{x})$ = $[\hat{f}(\bm{x}) - z_{1- \alpha /2}\sqrt{\hat{\sigma}_{\omega}^{2}(\bm{x}) + \hat{\sigma}^{2}(\bm{x})},\;\hat{f}(\bm{x}) + z_{1- \alpha /2}\sqrt{\hat{\sigma}_{\omega}^{2}(\bm{x}) + \hat{\sigma}^{2}(\bm{x})}]$\;

  \textbf{return:} CI$(\bm{x})$, PI$(\bm{x})$\;
   \caption{Pseudocode to obtain a confidence and prediction interval using concrete dropout.}
   \label{dropout}
\end{algorithm}

\subsection{A Toy Example with Homoscedastic Noise}
In this subsection, we demonstrate that a good on average performance on a single test set gives no guarantees for the actual quality of the uncertainty estimate. We simulated data from $y = f(x) + \epsilon$ with $f(x) = (2x-1)^{3}$, and $\epsilon \sim \N{0}{(0.2)^{2}}.$ The training and test set both contain $1000$ data points. The bootstrap method has an additional 150 validation data points to determine the data noise variance. The $x$-values are drawn uniformly from the interval $[-0.5, 0.5]$. A total of $M=50$ bootstrap networks are trained in order to construct a CI and PI for each $x$-value in the test set using the bootstrap approach. For the dropout approach, we used $B=100$ forward passes through the network. We repeated these procedures for estimating prediction and confidence intervals for 100 simulations of randomly drawn training sets.

\begin{figure}[t]
	\centering
   \subcaptionbox{$\alpha = 0.1$.}{\includegraphics[width=0.4\textwidth]{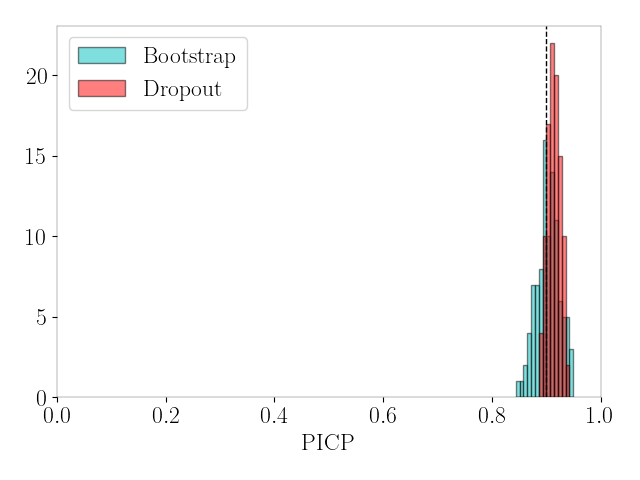}} %
  \subcaptionbox{ $\alpha = 0.2$.}{\includegraphics[width=0.4\textwidth]{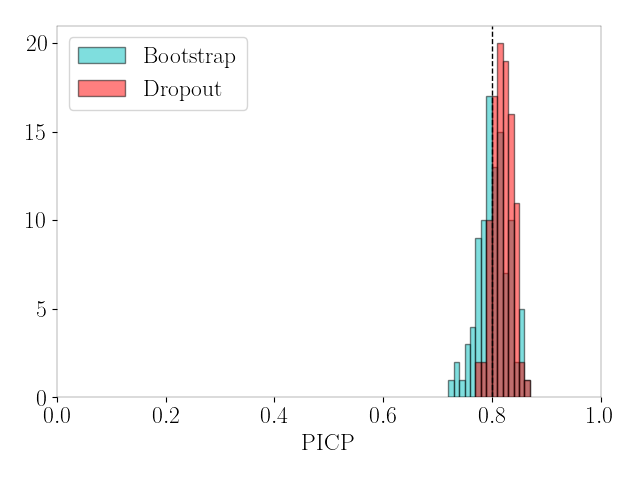}} 
  \caption{These histograms give 100 evaluations of the PICP, at a $(1-\alpha)$ confidence level using both the bootstrap and dropout approach. The PICP captures the fraction of data points in the test set for which the observations $y$ falls inside the corresponding prediction interval. The data is simulated from $y = f(x) + \epsilon$ with $f(x) = (2x-1)^{3}$, and $\epsilon \sim \N{0}{(0.2)^{2}}.$ The details of the construction of the PIs can be found in Section \ref{dropout and bootstrap}. From these histograms we can see that in a single simulation we would have a good performance of the PI on the test set with either method.}
\label{fig: PICPtoy}
\end{figure}

 In Figure \ref{fig: PICPtoy}, we see that both methods give a PICP that is very close to the desired values of 0.9 and 0.8 in each of the 100 simulations. Note that when simply using one data set, we would only have 1 PICP value. Furthermore, as we argued in Section \ref{whichestimate}, it is not clear that a good PICP indicates that the CIs are good. 

\begin{figure}[t!]
	\centering
   \subcaptionbox{$\alpha = 0.1$}{\includegraphics[width=0.4\textwidth]{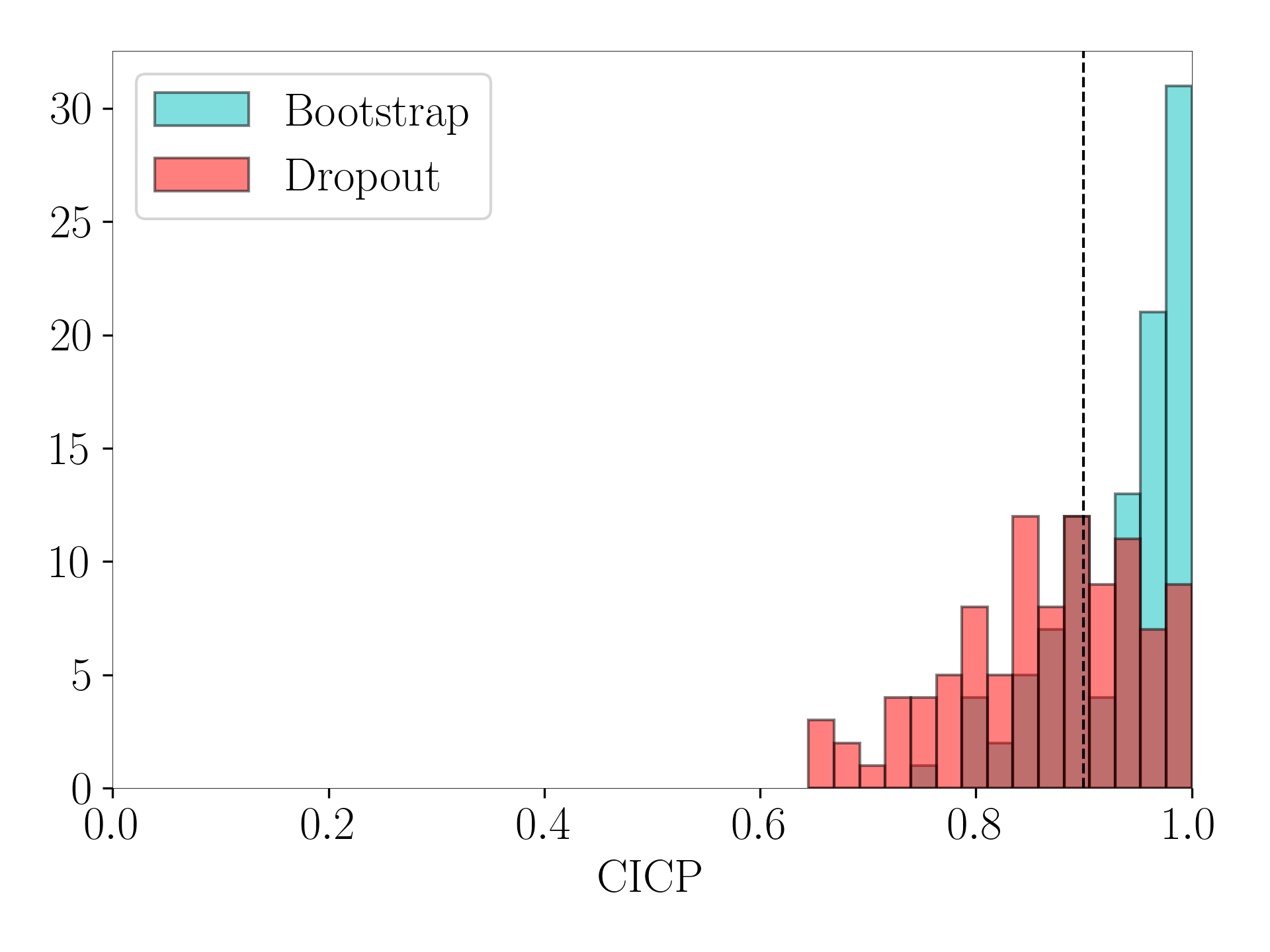}} 
  \subcaptionbox{$\alpha = 0.2$}{\includegraphics[width=0.4\textwidth]{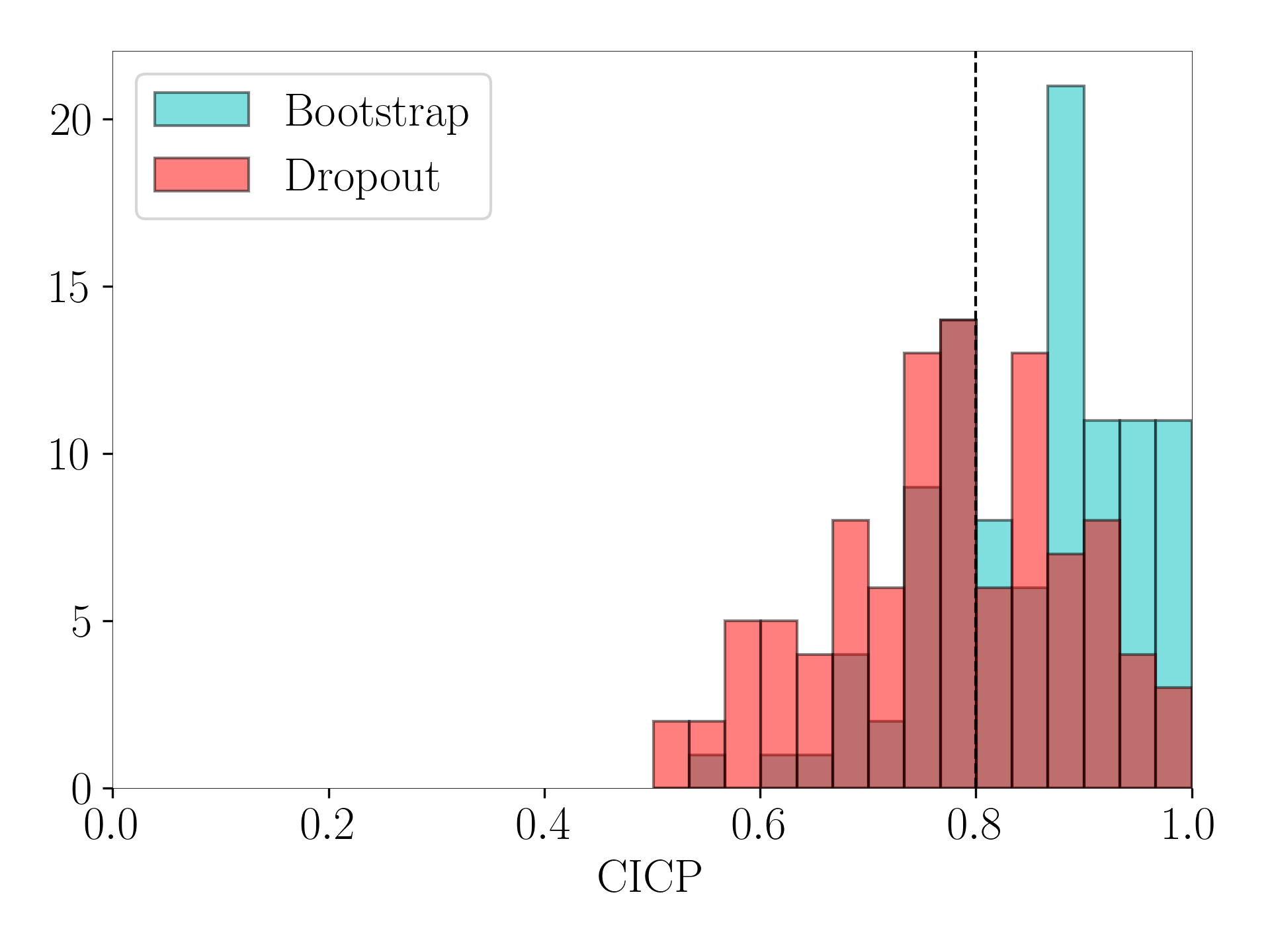}}\\
  \caption{These histograms give 100 evaluations of CICP, at a $(1-\alpha)$ confidence level. Each point in the histograms gives the fraction of datapoints in the test set of a new simulation for which the true function value $f(x)$ falls inside the corresponding confidence interval. Each simulation has its own CICP value. The same setup was used as in Figure \ref{fig: PICPtoy}. We observe that the good PICP values from Figure \ref{fig: PICPtoy} do not translate to good CICP values.}
\label{fig: CICPtoy}
\end{figure}

In a real-world scenario, we would not have access to the true function, $f(x)$, and we would not be able to calculate the CICP. In this case, however, we are. In Figure \ref{fig: CICPtoy}, we can see that the CICP values were not that great for most simulations. We see that there is a lot of variance between the CICP values of individual simulations. This exemplifies why using a single test set is not sufficient, as we argued in Section \ref{singleset}. Additionally, we observe that the confidence intervals of the bootstrap method are often too large, which we expected since the estimated model uncertainty uses the variance of an individual ensemble member and not of the average of the ensemble members. However, even \textit{if} these CICP values would have been perfect, it is still possible that this happens because the CIs are only correct on average. CIs that are much too large for some $x$ can be countered by CIs that are much too small for other $x$.

\begin{figure}[t!]
	\centering
   \subcaptionbox{PICF, $\alpha = 0.2$.}{\includegraphics[width=0.4\textwidth]{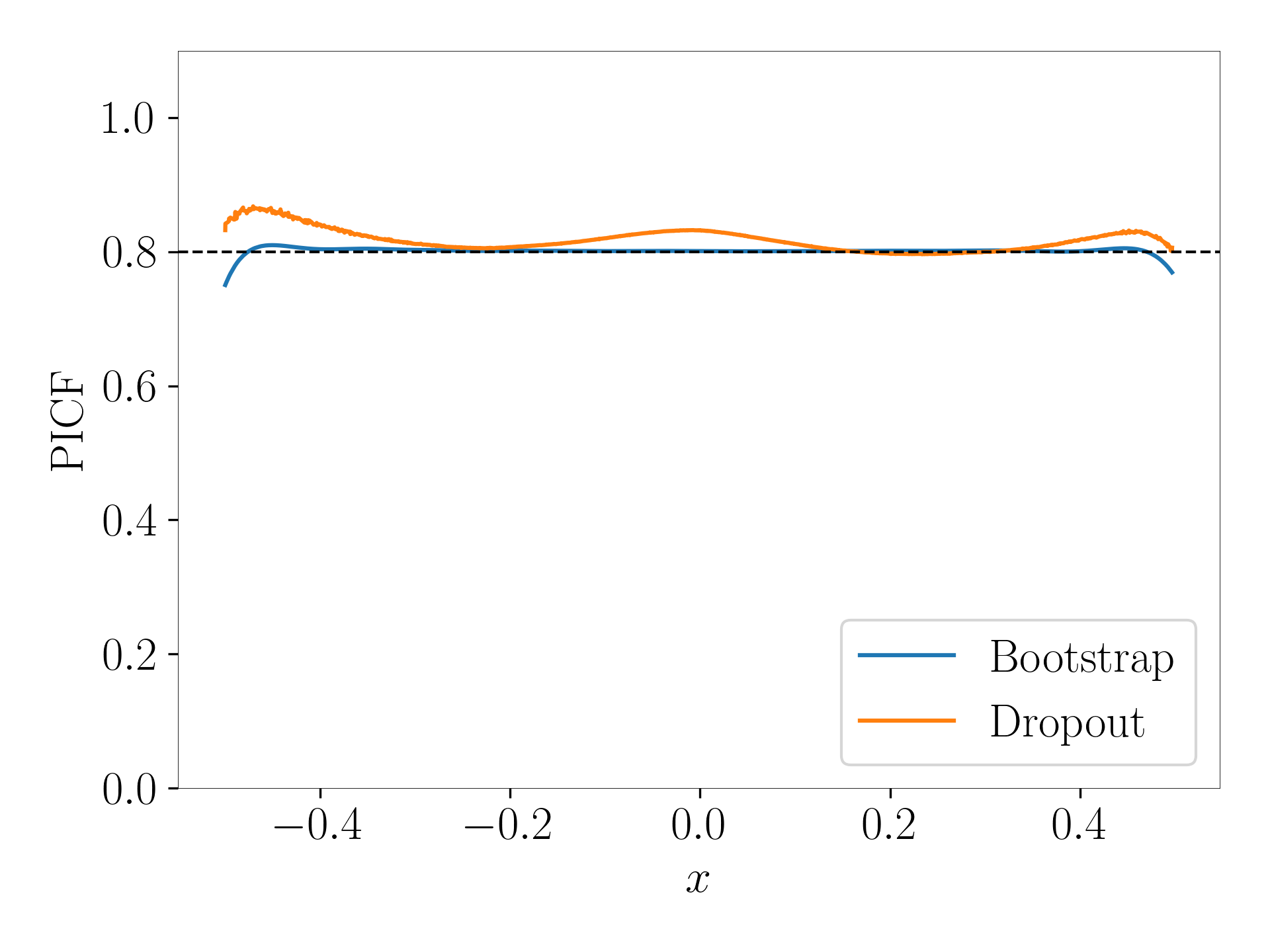}}
  \subcaptionbox{CICF, $\alpha = 0.2$.}{\includegraphics[width=0.4\textwidth]{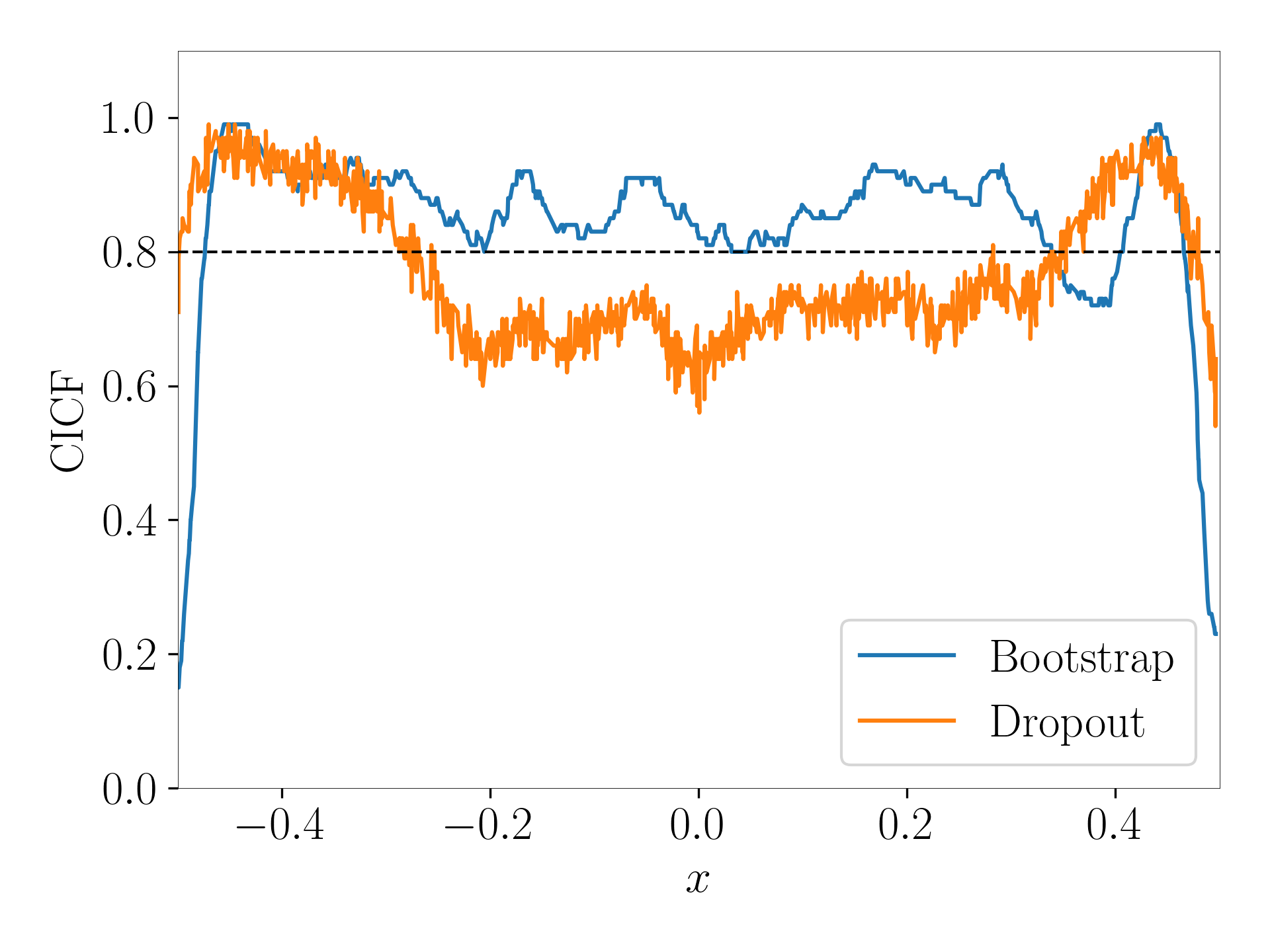}}
  \caption{The PICF and CICF plotted as a function of $x$. The PIs appear to be correct for most $x$ while the CIs are often too large or too small. The PIs and CIs were constructed at an $80\%$ confidence level. }
\label{fig: PICFCICFxtoy}
\end{figure}

This effect can be seen when we actually look at the coverage fraction per value of $x$ calculated over all the simulations, the CICF, as we proposed in the previous section. To reiterate, for the PICP and CICP we average over test data points and then provide a histogram over simulations, whereas for the PICF and CICF values we average over simulations. In Figure \ref{fig: PICFCICFxtoy}(b) we can see that for some values of $x$ the CIs contained the true value $f(x)$ in every simulation while hardly ever for other values of $x$. In Figure \ref{fig: PICFCICFxtoy}(a) we see that in this set of simulations the PIs are relatively accurate for most values of $x$ and not only on average.

\subsection{A Toy Example with Heteroscedastic Noise}
\begin{figure}[t]
   \centering
   \subcaptionbox{PICP, $\alpha=0.1$.}{\includegraphics[width=0.4\textwidth]{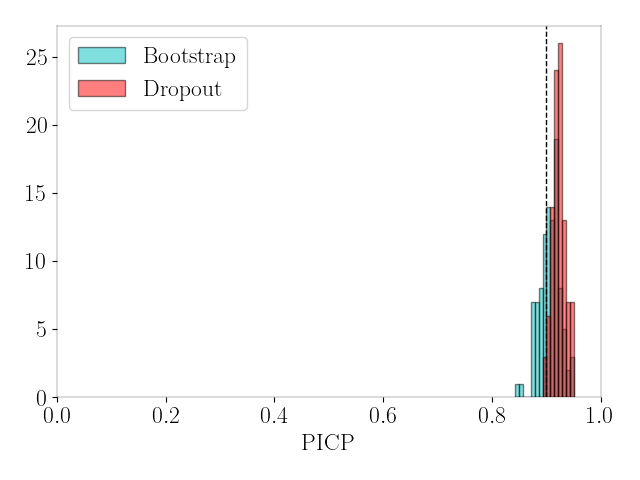}} 
  \subcaptionbox{PICF, $\alpha=0.1$. }{\includegraphics[width=0.4\textwidth]{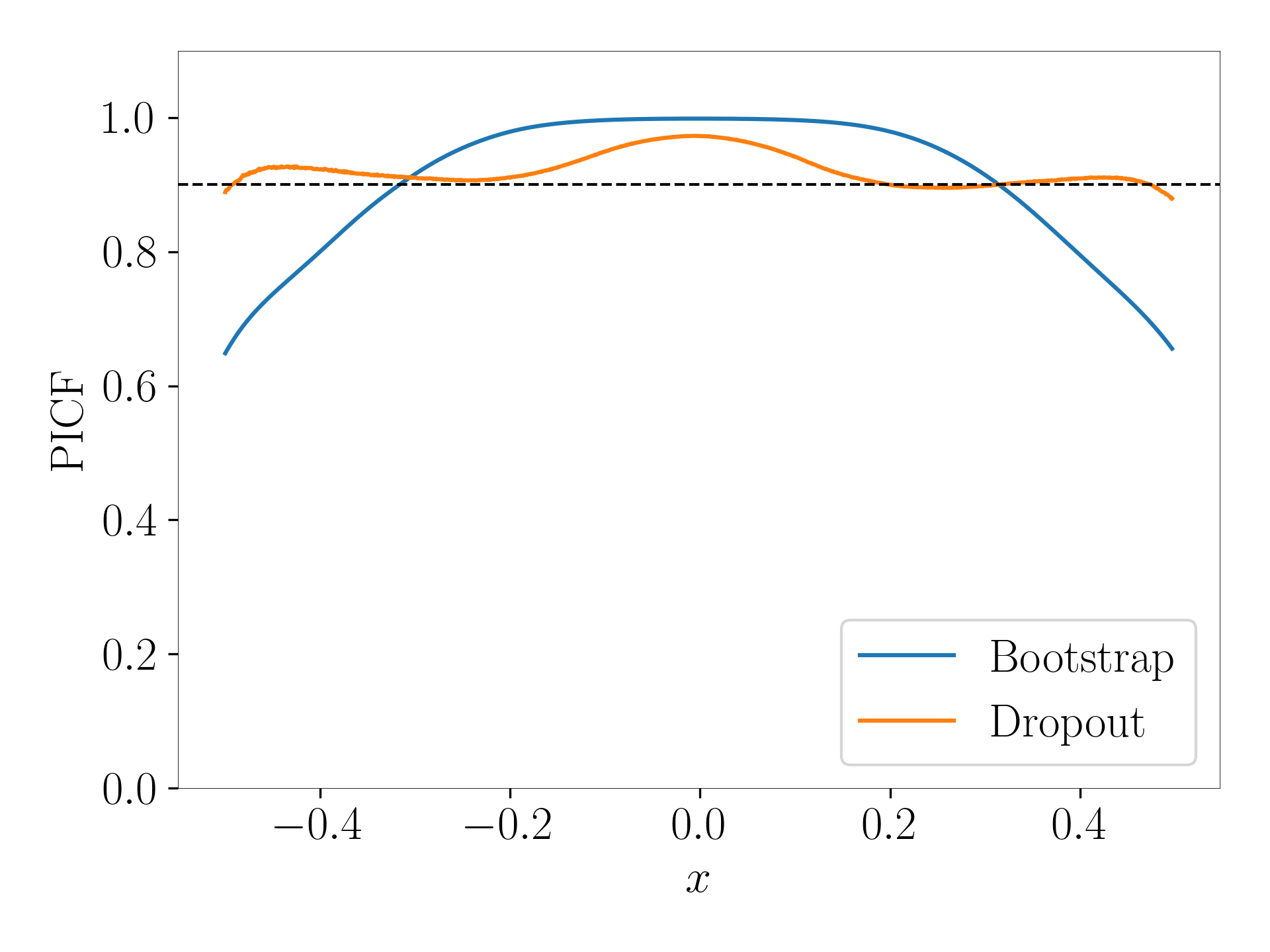}}
  \caption{These histograms give the different PICP and PICF values using the bootstrap and dropout approach. The PICP is obtained by calculating the coverage fraction of the PIs on the test set in each simulations. The PICF is obtained by calculating the coverage fraction of the PIs taken over all the simulations. In this simulation we constructed 90$\%$ PIs. Bootstrap has a Brier score of 0.011, dropout has a Brier score of 0.0011. We see that a good PICP score does not imply that the PIs are sensible for individual values of $x$. }
  \label{fig: PICPPICFhetero}
\end{figure}

In the previous example, the PIs behaved well. It is, however, also possible for the PICP to be good only because the PIs are correct on average. Figure \ref{fig: PICPPICFhetero} illustrates this point. We repeat the simulation but now using a noise term with a standard deviation of $0.1 + x^{2}$. The bootstrap method assumes homoscedastic noise, while concrete dropout does not. Figure \ref{fig: PICPPICFhetero} illustrates that the PICP score does not show us that the data noise variance estimate of the bootstrap method is wrong. Even worse, it can point in the wrong direction. We can favor the worse method if we would use the coverage fraction on a test set as our metric. Note that, in Figure \ref{fig: PICPPICFhetero}(a), we can see that, in most simulations, the bootstrap approach had a comparable or better PICP score compared to the dropout method. On average, the PICP score of the bootstrap method was even slightly closer to the desired value of 0.9. Figure \ref{fig: PICPPICFhetero}(b), however, shows that this is only the case because for some values of $x$ the coverage fraction was too high and for others too low, resulting in a good performance on average. According to the Brier score, dropout performed much better in this specific case.

\subsection{Out-of-Distribution Detection}
\begin{figure}[tb]
	\centering
   \subcaptionbox{Bootstrap}{\includegraphics[width=0.4\textwidth]{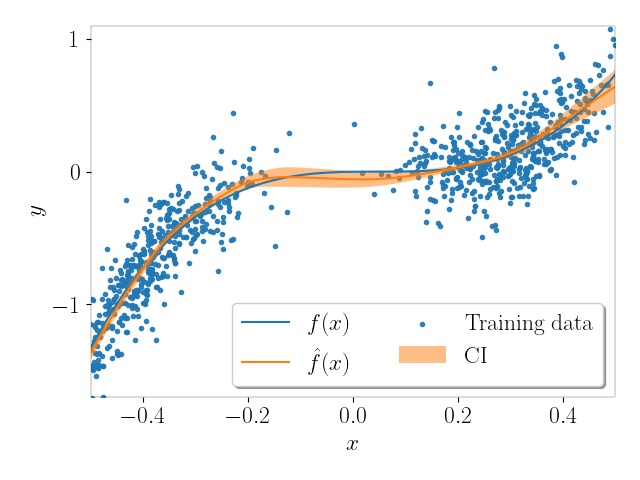}} 
  \subcaptionbox{Dropout}{\includegraphics[width=0.4\textwidth]{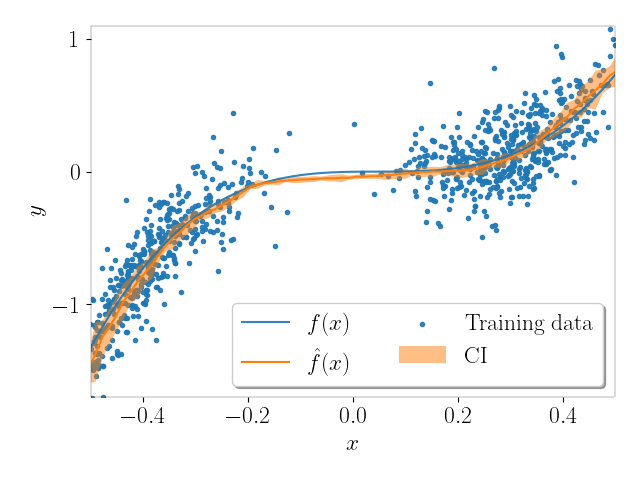}}
  \caption{These two figures give $80 \%$ CIs using the na\"ive bootstrap (a) and Monte Carlo dropout (b). The blue line is the true function and blue dots give the training data. The same function and noise were used as when making Figure \ref{fig: PICPtoy} with the difference that the covariates are not uniformly sampled. Even though the dropout approach gave a slightly better PICP score (0.83 versus 0.75) and higher average loglikelihood (0.14 versus 0.12), the CIs do not demonstrate better behaviour than those made with the bootstrap.}
\label{fig: OOD} 
\end{figure}

A desirable property of a confidence interval is that it gets larger in areas where there is a limited amount of data. It is not evident that a good performance on a test set guarantees this effect. In Figure \ref{fig: OOD}, we use the same function as in our example with homoscedastic noise, but simulate our $x$ values from a bimodal distribution instead of uniformly. Both models are trained using 1000 data points and are evaluated on a test set of size 1000. When making $80 \%$ CIs and PIs, the bootstrap and dropout methods give a PICP score of 0.75 and 0.83 respectively. The PICP score of the dropout method is closer to desired value of 0.8 and one might conclude from this that this method is better able to construct CIs. Additionally, the average loglikelihood was higher for the dropout method (0.14 versus 0.12). If we actually look at the CIs, however, we see that the behaviour is not as desired and that the bootstrap approach created more sensible CIs. The CIs of the bootstrap method get larger in the area around 0 where there is a limited amount of data and smaller around -0.4 and 0.3 where there is more data. The intervals created by using Monte Carlo dropout do the exact opposite, even though the performance on a single test set was better when using the PICP or loglikelihood as a metric.  

\begin{figure}[tb]
	\centering
   \subcaptionbox{PICP}{\includegraphics[width=0.4\textwidth]{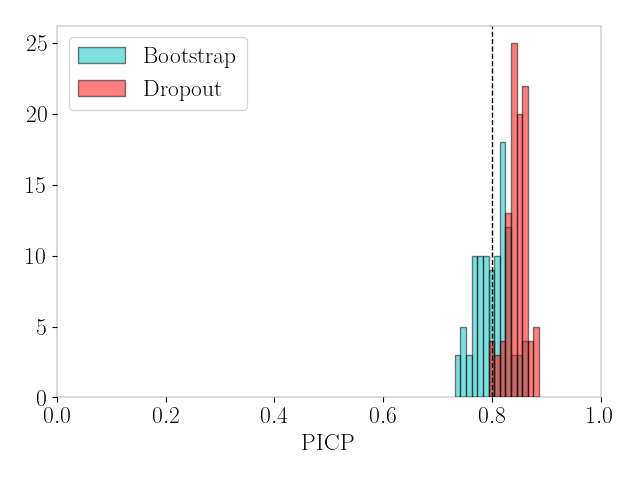}}
  \subcaptionbox{CICF}{\includegraphics[width=0.4\textwidth]{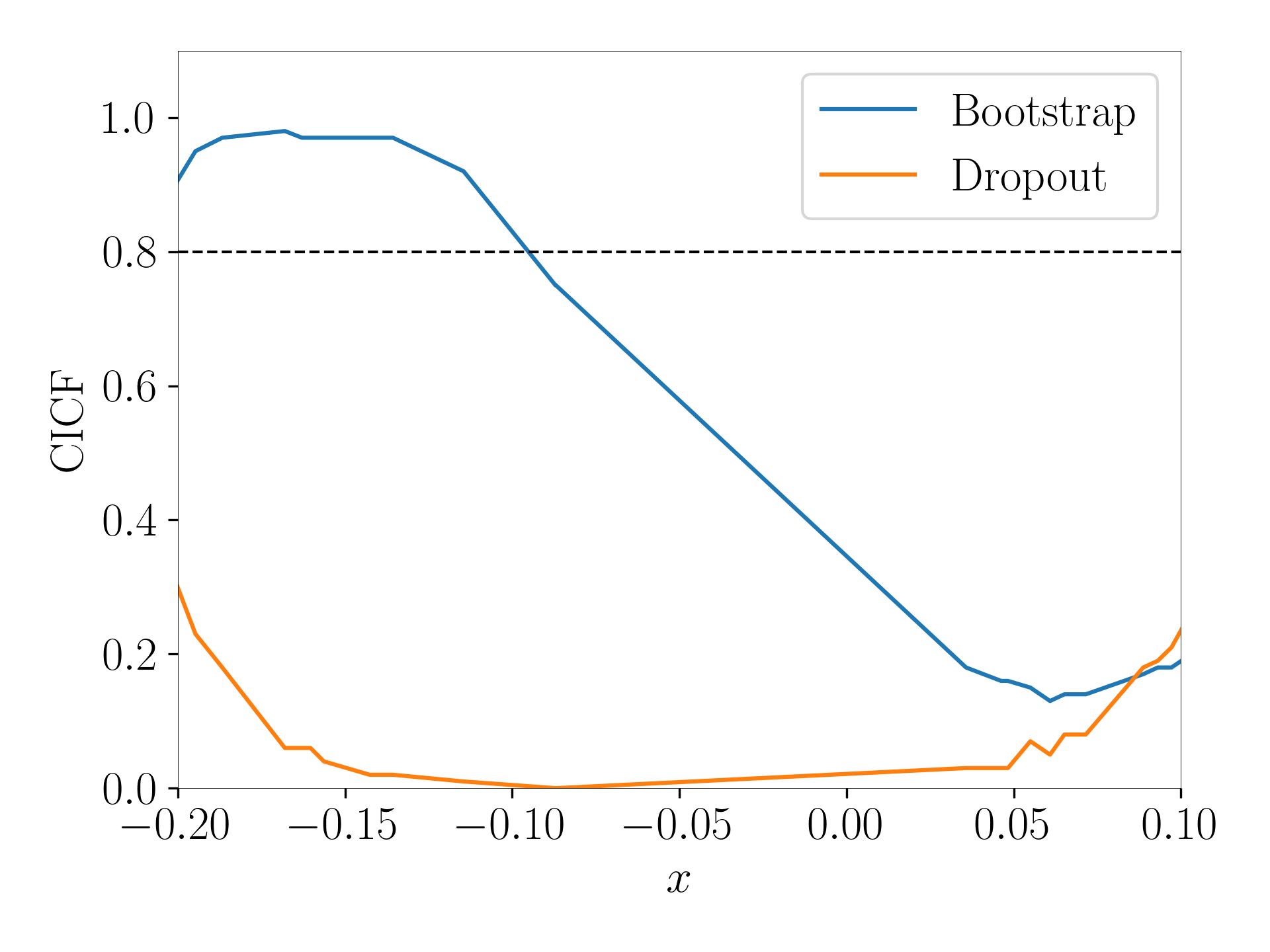}} \hspace{1em}
  \caption{This figure demonstrates that a good PICP value (a) does not guarantee desirable behaviour of the model uncertainty estimates. The region between -0.2 and 0.1 contained fewer data points. This figure, combined with Figure \ref{fig: OOD}, demonstrates that both methods behave very differently in areas of limited data, something that is not detectable by merely evaluating the predictive uncertainty on a test set. }
\label{fig: OODsimulation}
\end{figure}

We simulated the data a total of 100 times to further demonstrate this behaviour. We can see in Figure \ref{fig: OODsimulation}(a) that in all 100 simulations both methods got a reasonable PICP score. If we look, however, at the CICF for $x$ values between $-0.2$ and $0.1$ we notice that dropout was not able to determine the uncertainty accurately\footnote{We suspect that the behaviour of dropout is a result of the interpolation. When extrapolating, the ReLu activation functions cause the function values to increase. This gives rise to a large variance in the forward passes through the network. In the region around $x=0$, the function values are almost zero, likely resulting in less variance and thus a smaller confidence interval. This explanation ignores subtleties with bias terms and it may be interesting to investigate this type of behaviour of dropout further.}. Additionally, both methods had a substantial bias in the area with fewer data points. This information is only available if we use simulated data where the true function is known. Since both methods assume that the model is unbiased, it is unsurprising that the coverage is not perfect.  As a side-note, some other methods, such as \citet{zhou2018approximation} do explicitly take this into account.

\subsection{Boston Housing} \label{Boston Housing}

So far, we only used a simple one-dimensional simulation in our testing procedure. A good testing procedure should: 1 - be representative for real-world problems, and 2 - allow easy comparison between different methods. The data sets that are currently being used, listed in Table \ref{Tab: datasets}, meet these criteria. As we just demonstrated, however, it is desirable to simulate the data to get more insights in the accuracy of the uncertainty estimates. A solution would be to set up simulations based on data sets that are currently being used. In the following algorithm we give a suggestion how we could set up a simulation that resembles the Boston Housing data set. 

\begin{algorithm}[tb]
\textbf{Inputs} A real world data set $\mathcal{D} = \{(\bm{x}_{1}, y_{1}), \ldots (\bm{x}_{N}, y_{N})\}$\;
  Train a random forest on $\mathcal{D}$ and use this predictor as the true function $f(\bm{x})$\;
  Calculate the residuals, $(y_{i} - f(\bm{x}_{i}))$\;
  Train a second random forest that predicts the residuals squared as a function of $\bm{x}$ and use this predictor as the true variance $\sigma^{2}(\bm{x})$\;
  Simulate a new data set  $\mathcal{D}_{\text{new}} = \{(\tilde{\bm{x}}_{1}, \tilde{y}_{1}), \ldots (\tilde{\bm{x}}_{N}, \tilde{y}_{N})\}$, where $\tilde{\bm{x}}_{i} = \bm{x}_{i}, \quad$ and $\quad \tilde{y}_{i} \sim \N{f(\tilde{\bm{x}}_{i})}{\sigma^{2}(\tilde{\bm{x}}_{i})}$ \;
  \textbf{Return:} $\mathcal{D}_{\text{new}}$\;
   \caption{Pseudo-code to simulate data based on the Boston Housing data set}
\end{algorithm}

\begin{figure*}[tb]
	\centering
   \subcaptionbox{PICP}{\includegraphics[width=8.5cm]{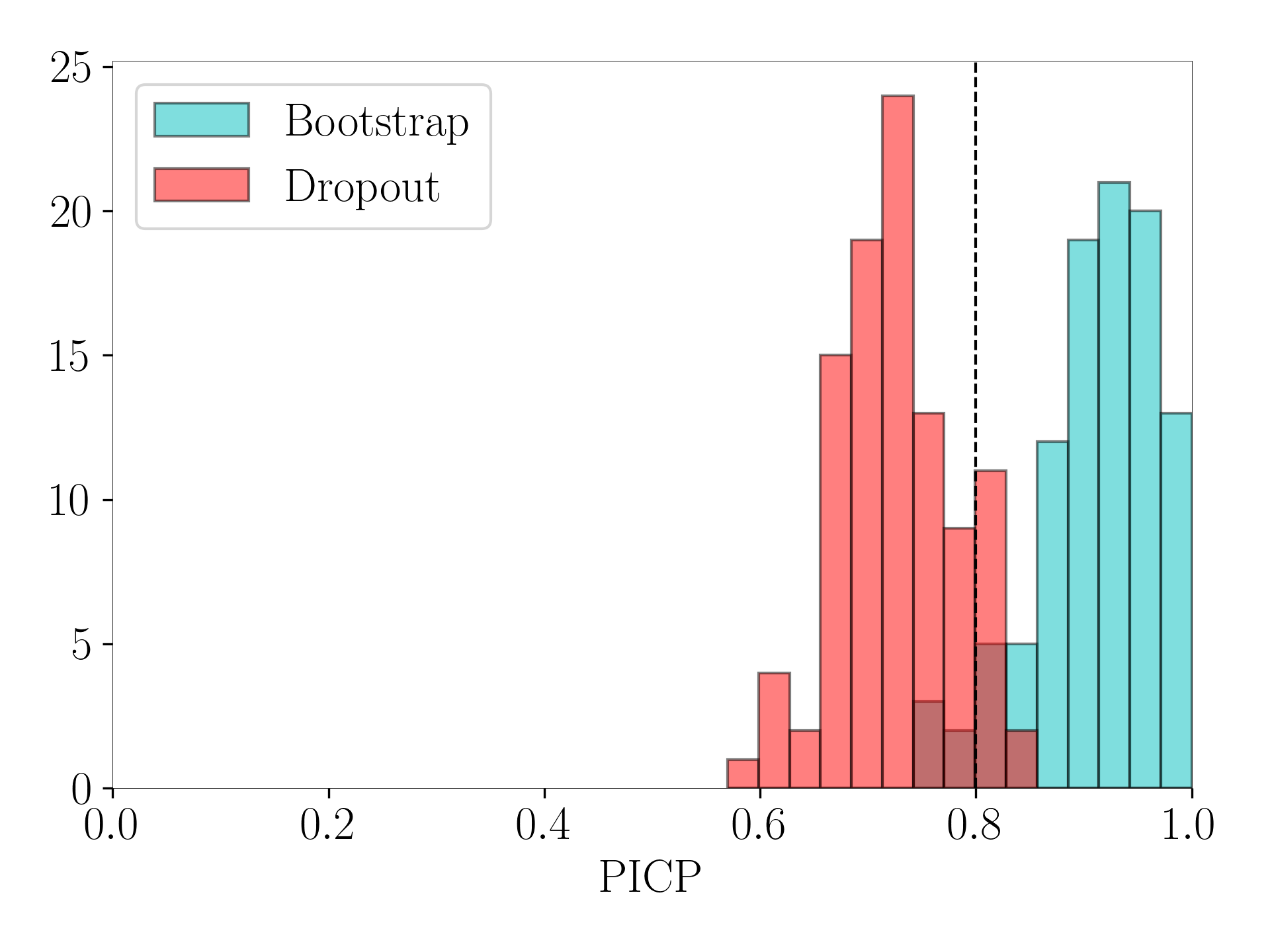}}  
  \subcaptionbox{PICF}{\includegraphics[width=8.5cm]{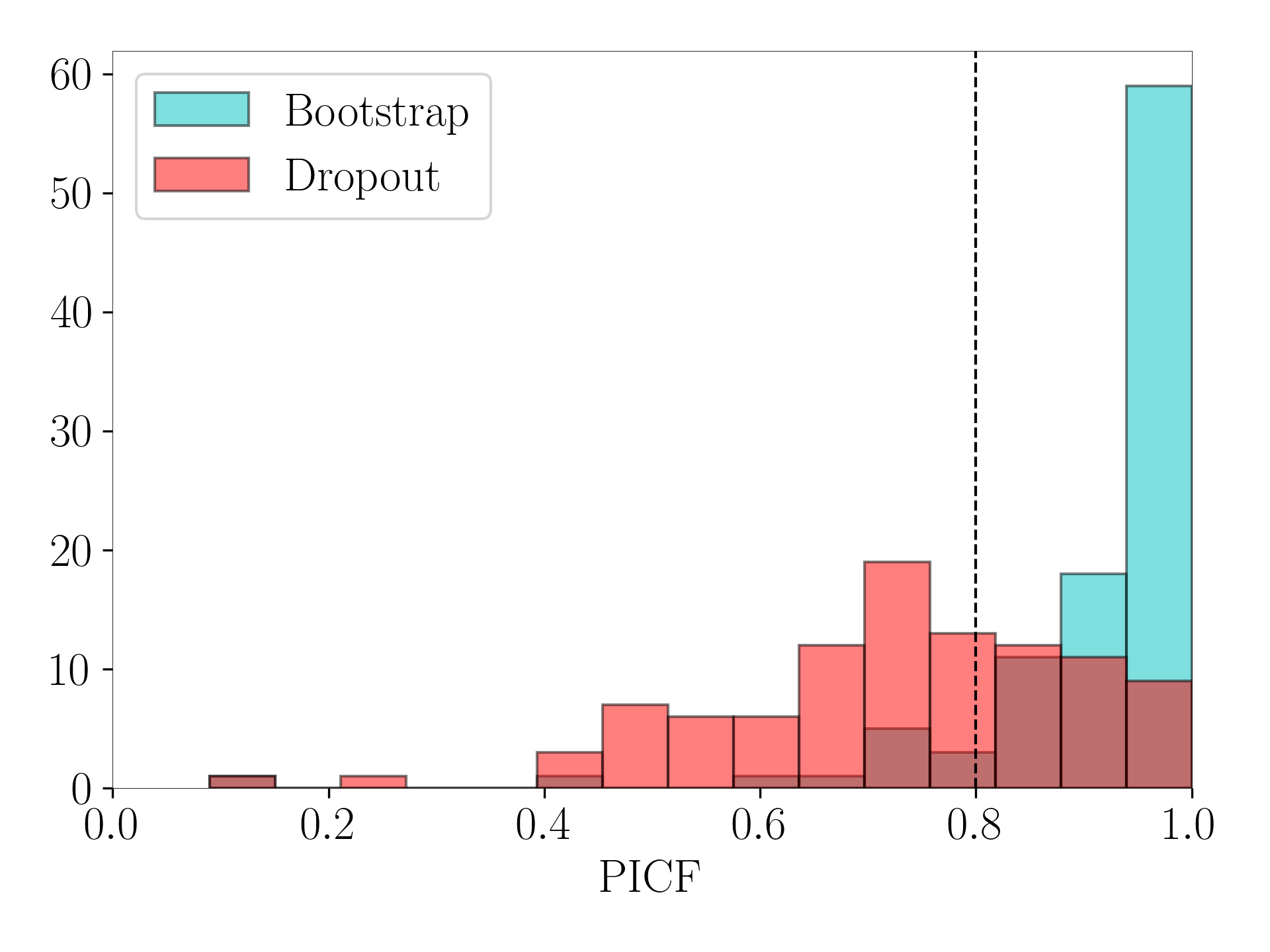}}  \\
  \subcaptionbox{CICP}{\includegraphics[width=8.5cm]{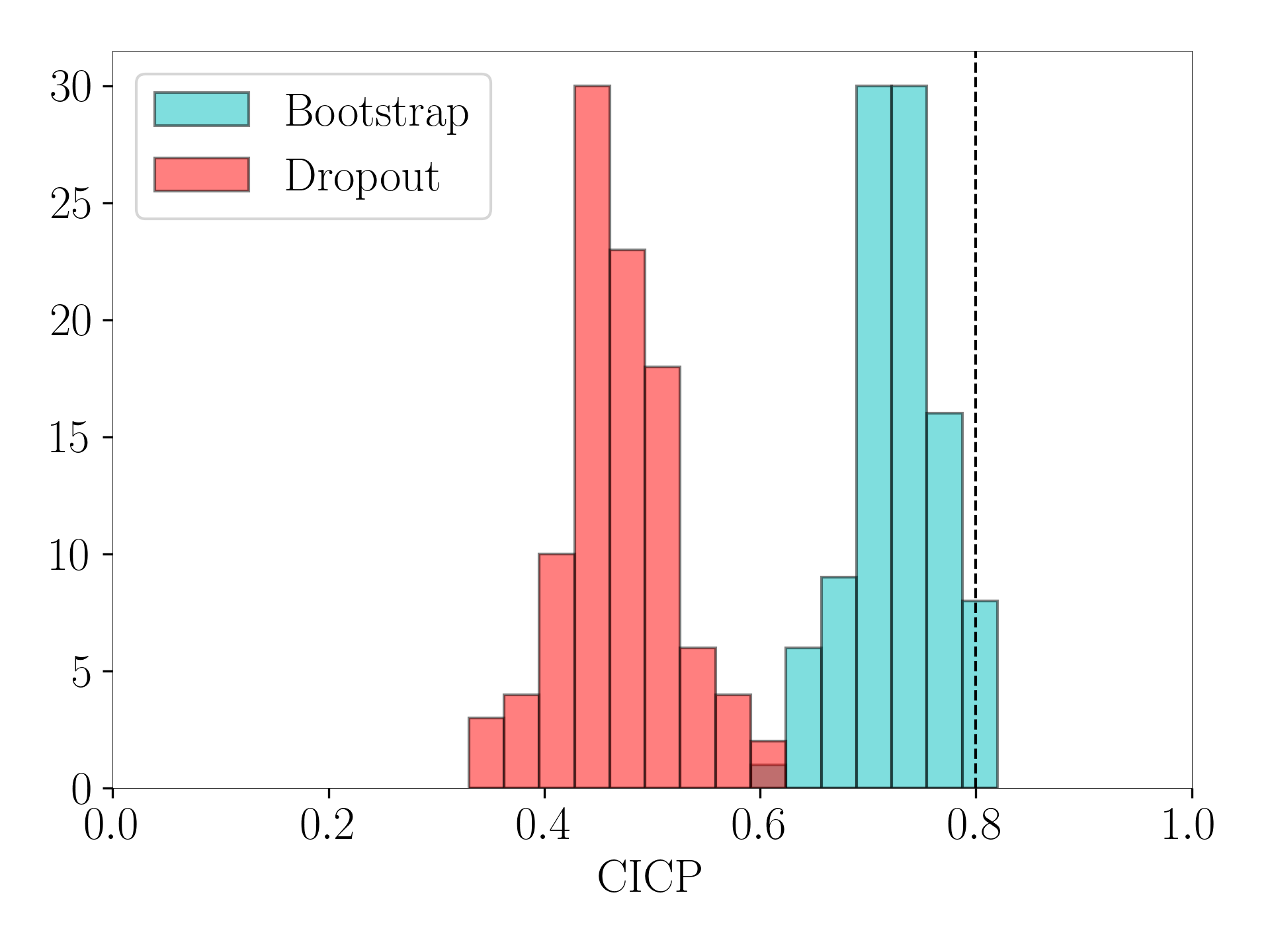}}  
  \subcaptionbox{CICF. }{\includegraphics[width=8.5cm]{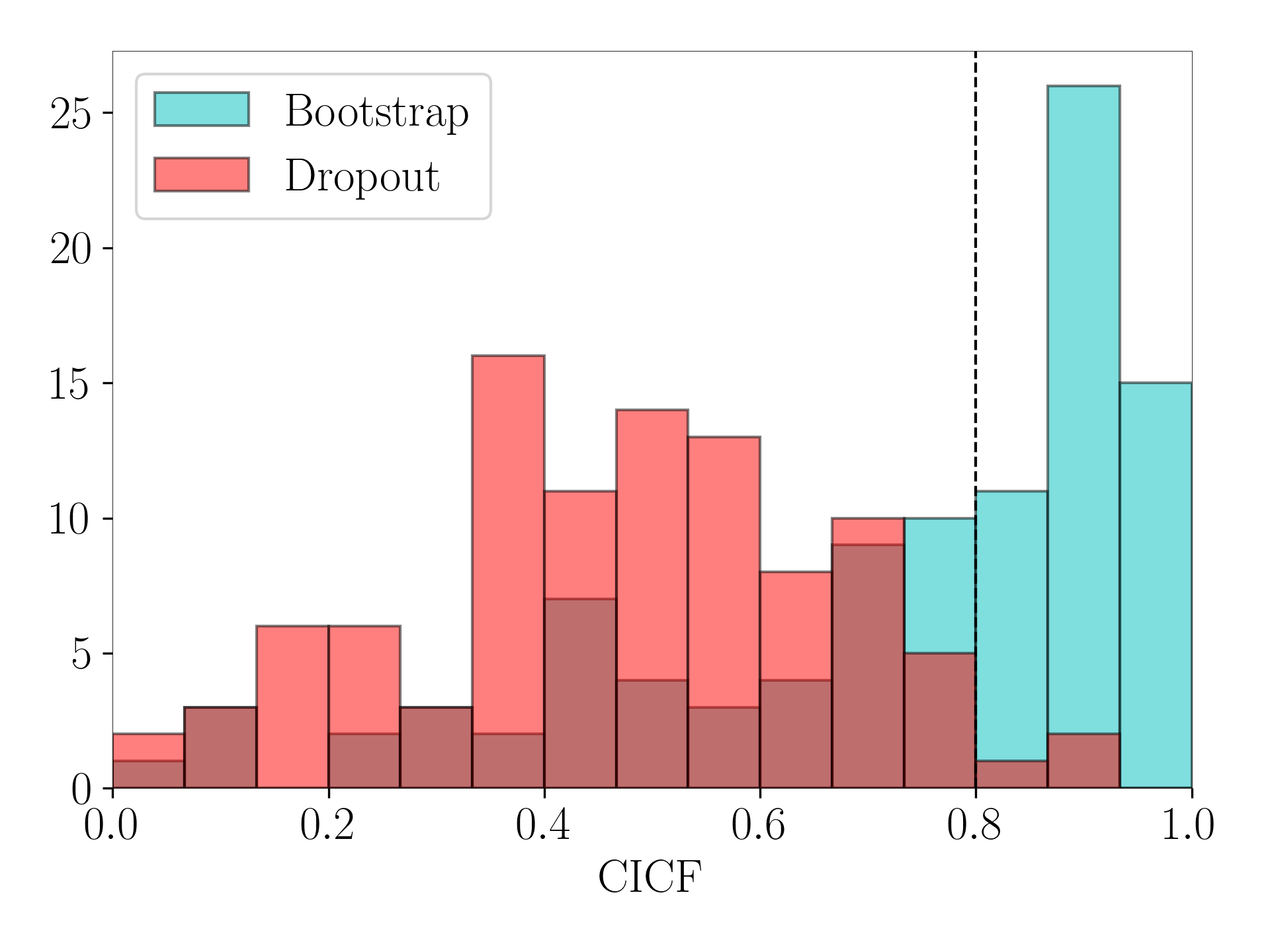}} 
  \caption{These histograms give the different PICP, PICF, CICP, and CICF values using the bootstrap and dropout approach on the Boston Housing simulation. The PICP and CICP values are obtained by calculating the coverage fraction of the PIs and CIs on the test set in each simulation. The PICF and CICF values are obtained by calculating the coverage fraction of the PIs and CIs for each test data point taken over all the simulations. In this simulation we constructed 80$\%$ PIs and CIs. For the PICF, bootstrap has a Brier score of 0.027 and average width of 14.2. Dropout has a Brier score of 0.032 and an average width of 8.3. For the CICF, bootstrap has a Brier score of 0.147 and average width of 4.92. Dropout has a Brier score of 0.15 and average width of 3.04. We once more observe that a (relatively) good PICP value gives no guarantees for the actual performance of either the PI or CI on individual data points.}
\label{fig: PICPPICFboston}
\end{figure*}

As an illustration, we implemented this idea using two random forests with 100 trees, and max depth 15. These hyperparameters resulted from a manual grid search. We do not attempt to simulate new $\bm{x}$-values as it would be difficult to get the dependencies between the covariates correct. We divided the generated data set in a train, test, and validation set of sizes 366, 100, and 40. The validation set is used by the bootstrap to determine the estimate of the data noise variance. The same bootstrap and dropout procedures as in the previous subsections were used to obtain PIs and CIs. In Figure \ref{fig: PICPPICFboston}(a), we can see that the PICP was a little too high on average for the bootstrap method a little too low for the dropout method but quite close to the desired value of 0.8 in most of the simulations. In Figure \ref{fig: PICPPICFboston}(b), we can see, however, that for almost every $\bm{x}$, the prediction intervals were either too large or too small. We see a similar trend if we look at the CICP in Figure \ref{fig: PICPPICFboston}(c). Both methods consistently had a CICP that was too low. This enforces the argument that simply looking at the performance on a single test set is far from optimal. It also shows that it is possible to apply this simulation-based testing procedure to representative data sets.

\section{CONCLUSION} \label{Conclusion}
We conclude that the testing methodology applied to many recent publications for evaluating the quality of uncertainty estimates leaves a lot of room for improvement, especially if the eventual application of a method is the construction of a confidence or prediction interval. Both the loglikelihood and the PICP evaluate the predictive uncertainty, which is a combination of the data noise variance and model uncertainty, on a previously unseen test set. A good predictive uncertainty overall, however, is not necessarily indicative of a good estimate of the model uncertainty or data noise variance. Since the true function values are unknown, it is impossible to test confidence intervals directly. For the loglikelihood, we demonstrated that a better score does not guarantee better prediction intervals. Additionally, it is not clear how to compare methods that output a density with methods that output a prediction or confidence interval directly. Furthermore, we showed that the PICP score does not test coverage correctly and at best measures marginal coverage. A stronger and more useful characteristic is correct pointwise coverage. 

To overcome these problems, we propose simulation-based testing to evaluate pointwise coverage of the prediction and confidence intervals. We note that we assume that the eventual application of the method is to give accompanying prediction or confidence intervals. In order to properly test coverage, it is necessary to repeat the experiment: create a new data set, train the model, create new intervals. Possible quantitative metrics of the PIs and CIs are the Brier score of the PICF/CICF and the average width of the intervals. This approach has some downsides. The computational demands for running these tests are higher and there is a need to simulate the data. We propose to set up simulations based on the data sets listed in Table \ref{Tab: datasets}. It is also possible to set up a simulation based on a data set of particular interest. The additional computational demands only play a role during the testing of these uncertainty quantification methods and not in their usage in practice. 
\vskip 0.2in
\subsection{Future Research}
In order to compare different uncertainty estimation methods, it is necessary to use the same simulations. It would therefore be useful to create a number of benchmark simulations that can be used to test uncertainty estimates. We propose to base these simulations on the data sets in Table \ref{Tab: datasets}. In Section \ref{Boston Housing} we gave a simple demonstration of such a simulation. In this paper, we only considered normally distributed noise. Different distributions would allow us to explicitly see what happens if the customary assumption of normality does not hold. The methods in this paper are based on this assumption but a method like quantile regression, for instance, is not. 

We saw in Section \ref{Experimental} that a method that performs better on a test set need not have better behaving uncertainty estimates. With new benchmarks, it would be worthwhile to re-evaluate currently available methods for estimating uncertainty.

\FloatBarrier
\section*{References}
\bibliographystyle{apalike}
\bibliography{../../../references3}

\end{document}